\documentclass[final,5p,times,twocolumn]{elsarticle}

\usepackage{amssymb}
\usepackage{amsmath}
\usepackage{amsthm}
\usepackage{textcomp}

\journal{Reliability Engineering \& System Safety}

\usepackage{multirow}
\usepackage{pdflscape}

\usepackage[table]{xcolor}
\usepackage{subcaption}
\definecolor{lightgray}{gray}{0.4}

\usepackage{xurl}

\usepackage{array}
\usepackage{ltablex}
\keepXColumns
\usepackage{ragged2e} 

\usepackage{booktabs}    
\usepackage{array} 

\begin{document}
\begin{frontmatter}
\title{Hazard Management in Robot-Assisted Mammography Support}



\author[aff1]{Ioannis Stefanakos\corref{cor1}}
\ead{ioannis.stefanakos@york.ac.uk}

\author[aff2]{Roisin Bradley}

\author[aff1]{Radu Calinescu}
\ead{radu.calinescu@york.ac.uk}

\author[aff3]{Beverley Townsend}
\ead{bev.townsend@york.ac.uk}

\author[aff1]{Tianyuan Wang}
\ead{tianyuan.wang@york.ac.uk}

\author[aff4]{Jihong Zhu}
\ead{jihong.zhu@york.ac.uk}

\cortext[cor1]{Corresponding author}

\affiliation[aff1]{organization={Department of Computer Science},
            addressline={University of York},
            city={York},
            country={United Kingdom}}

\affiliation[aff2]{organization={York and Scarborough Teaching Hospitals NHS Foundations Trust},
            city={York},
            country={United Kingdom}}

\affiliation[aff3]{organization={York Law School},
            addressline={University of York},
            city={York},
            country={United Kingdom}}

\affiliation[aff4]{organization={School of Physics, Engineering and Technology},
            addressline={University of York},
            city={York},
            country={United Kingdom}}

\begin{abstract}
Robotic and embodied-AI systems have the potential to improve accessibility and quality of care in clinical settings, but their deployment in close physical contact with vulnerable patients introduces significant safety risks. This paper presents a hazard management methodology for MammoBot, an assistive robotic system designed to support patients during X-ray mammography. To ensure safety from early development stages, we combine stakeholder-guided process modelling with Software Hazard Analysis and Resolution in Design (SHARD) and System-Theoretic Process Analysis (STPA). The robot-assisted workflow is defined collaboratively with clinicians, roboticists, and patient representatives to capture key human–robot interactions. SHARD is applied to identify technical and procedural deviations, while STPA is used to analyse unsafe control actions arising from user interaction. The results show that many hazards arise not from component failures, but from timing mismatches, premature actions, and misinterpretation of system state. These hazards are translated into refined and additional safety requirements that constrain system behaviour and reduce reliance on correct human timing or interpretation alone. The work demonstrates a structured and traceable approach to safety-driven design with potential applicability to assistive robotic systems in clinical environments.
\end{abstract}


\begin{keyword}
assistive and medical robotics\sep human–robot interaction\sep hazard analysis\sep safety-critical systems\sep mammography


\end{keyword}

\end{frontmatter}


\section{Introduction}

The landscape of artificial intelligence is currently undergoing a transformative shift from ``disembodied'' algorithms to \emph{embodied AI}---systems where intelligence is integrated into physical robotic bodies capable of perceiving, moving within, and interacting with the real world. Unlike traditional AI, which processes data in isolation, embodied AI agents leverage advanced sensorimotor coordination to perform complex physical tasks with high precision. This technological leap has taken robotics from controlled industrial settings into collaborative human environments, allowing for sophisticated human-robot interaction (HRI) that prioritises safety and adaptability in dynamic environments.

In assistive care, embodied-AI provides significant potential to alleviate mounting pressures on global healthcare infrastructure and clinical labour shortages. In the domain of physical assistance, these systems are already being utilised for tasks such as robotic-assisted patient transfer~\cite{burkman2017further}, automated limb rehabilitation  for stroke recovery~\cite{hodkin2018automated}, interactive dressing assistance~\cite{zhu2024you} and assistance with meal preparation~\cite{doi:10.1177/02783649261420234}. Central to these applications is the ability to manage direct physical and force interactions while accurately estimating human posture and intention. Mastering these elements is vital to ensuring that robotic interventions are both safe and responsive to user movements and changing needs, particularly in clinical procedures that require high-fidelity physical support.

\vspace*{-0.5mm}
One such critical application is X-ray mammography, the global standard for early breast-cancer detection~\cite{cdc-2025,Marmot2013,PHE2023guidance}. While highly effective, the procedure is physically demanding, requiring patients to maintain strenuous, stable positions that are essential for the acquisition of diagnostic-quality images. This creates a significant care gap for individuals who cannot meet these positioning requirements due to physical impairments, systematically excluding a vulnerable population from life-saving breast screening. As radiographers cannot provide physical support during X-ray imaging because of radiation exposure risks, there is a critical need for a robotic-assistive solution capable of ensuring stable, compliant positioning for these individuals. To address this need, we are developing MammoBot~\cite{cruk-2024,bcn-2026}, a dual-arm assistive robotic system designed to provide X-ray mammography support to patients with reduced upper-body strength.

\begin{figure*}[!t]
    \centering
    \begin{subfigure}[b]{0.46\textwidth} 
        \centering
        \includegraphics[width=\textwidth]{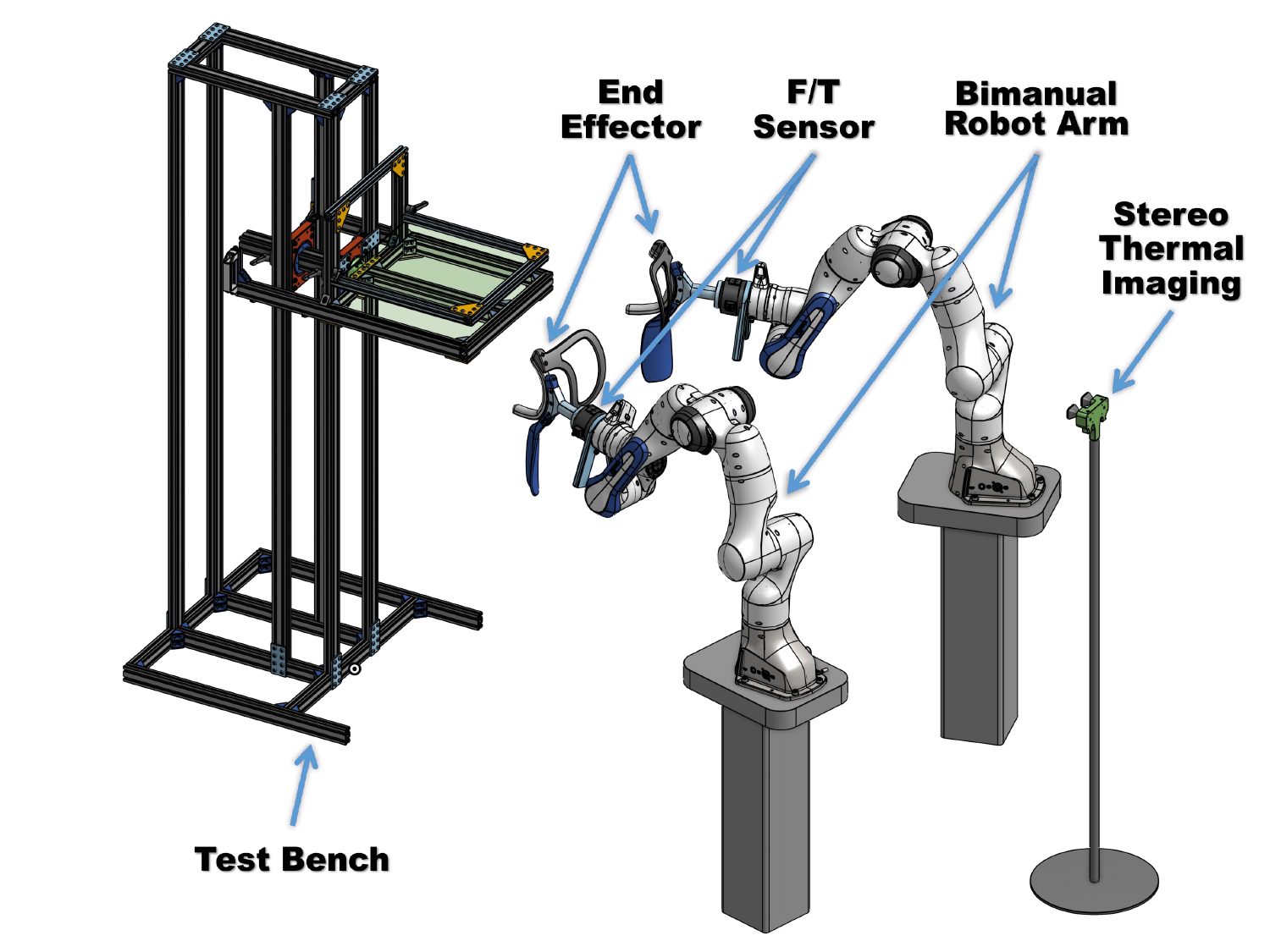}
        \caption{}
        \label{fig:full_assem}
    \end{subfigure}
    \begin{subfigure}[b]{0.52\textwidth} 
        \centering
        \includegraphics[width=\textwidth]{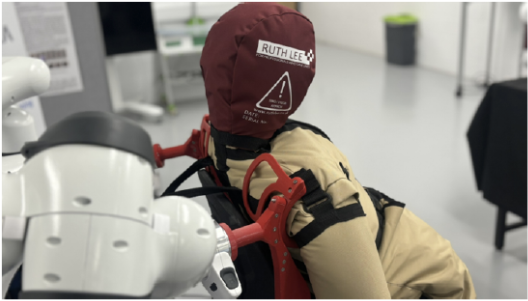}
        \caption{}
        \label{fig:system_in_action}
    \end{subfigure}
    \caption{The MammoBot bimanual assistive system. (\textbf{a}) System overview depicting an X-ray mammography machine mock-up we used for testing purposes, the bimanual robotic manipulation platform with end-effectors, and the thermal imaging system. (\textbf{b}) Proof-of-concept 3D-printed end-effectors we co-designed with York Hospital (UK) radiographers and radiologists (red), fitted to two Franka Research~3 robotic arms. Tests conducted at our York Institute for Safe Autonomy lab on a 50kg human-weighted mannequin equipped with force/torque sensors confirmed the need for a bimanual configuration to safely support patients in the required mammography positions, and determined the necessary robotic-arm payload range (8--14~kg, depending on user weight and required level of support).}
    \label{fig:overall_system}
\end{figure*}

However, the close physical interaction between the MammoBot system and vulnerable patients introduces significant safety risks. The hazards responsible for these risks often emerge less from component failures, and more from the complex interactions between the patient, the radiographer, and the robot controllers. This paper introduces a systematic hazard management methodology developed during the initial stages of our MammoBot project and designed to mitigate these safety risks. Our methodology integrates: (1)~stakeholder-guided process modelling; (2)~hazard identification through a combination of established safety-analysis techniques, i.e., Software Hazard Analysis and Resolution in Design (SHARD)~\cite{521885}, and Systems-Theoretic Process Analysis (STPA)~\cite{10.7551/mitpress}; and (3)~stakeholder-guided refinement of the initial system design. This provides a structured and traceable framework for the safety-driven design of clinical robot-assistive systems. By identifying and mitigating safety risks during early system-development stages, our methodology ensures that safety requirements are integrated into vulnerable-human–robot collaboration processes from the outset.

The remainder of the paper is structured as follows. Section~\ref{sec:system} introduces the MammoBot system and architecture, and describes an example clinical scenario illustrating its intended use. Section~\ref{sec:background} provides background on hazard analysis techniques for assessing safety in socio-technical systems. Section~\ref{sec:related_work} reviews related work on safety in collaborative and assistive robotics, hazard analysis methodologies, and regulatory frameworks. Section~\ref{sec:methodology} presents our hazard management methodology, including the collaborative process design and the application of complementary hazard analysis techniques to identify technical and procedural deviations and analyse unsafe human–system interactions, followed by refinement of system requirements based on the findings. Section~\ref{sec:discussion} discusses the implications of the hazard analyses and the challenges of designing safe assistive robotic systems for use in clinical environments. Finally, Section~\ref{sec:conclusion} concludes the paper with a brief summary, and outlines directions for future work.

\section{The MammoBot system\label{sec:system}}

\subsection{Motivation}

The MammoBot project is motivated by the need to address a fundamental inequity in breast cancer screening. Breast cancer remains a leading cause of mortality globally, with early detection through routine X-ray mammography saving thousands of lives annually \cite{cdc-2025, sung2021global, NHS2021}. However, the physical positioning requirements of the procedure act as a barrier for patients with reduced upper-body strength. In the UK alone, it is estimated that tens of thousands of eligible women, including many wheelchair users and other individuals with restricted mobility, face substantial barriers to accessing this service~\cite{NYBSS2023,PHE2018inequalities}. Furthermore, screening-centre staff cannot provide the necessary physical support because of the high levels of radiation that that would entail. Our MammobBot system~\cite{cruk-2024,bcn-2026} is being developed to enable women with such physical impairments to achieve and maintain the positioning required for diagnostic-quality mammography images.

\subsection{Architecture}

The system (depicted in Figure~\ref{fig:overall_system}) is a dual-arm assistive robotic platform engineered to provide stable and adaptive physical support to patients during mammography. Its architecture is designed around the principles of safety, patient privacy, and clinical efficacy. It consists of three primary subsystems: a robotic manipulation platform, a non-visual perception module for patient tracking, and bespoke end-effectors for patient contact. We describe each of these subsystems below.

\begin{figure*}[!t]
    \centering
    \begin{subfigure}[b]{0.48\textwidth} 
        \centering
        \includegraphics[trim=0mm 0mm 350mm 0mm, clip, width=0.68\textwidth]{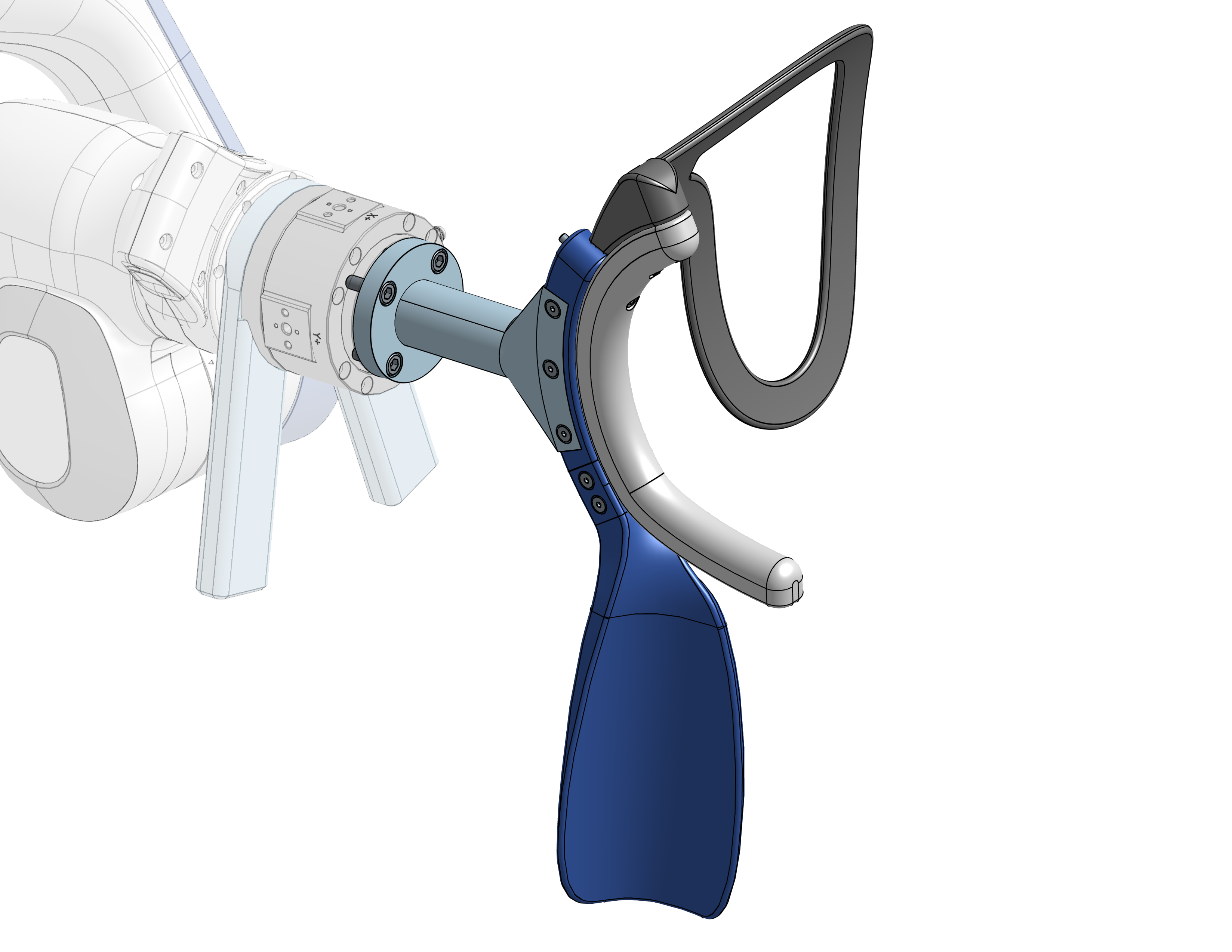}
        \caption{}
        \label{fig:tool_closed}
    \end{subfigure}
    \begin{subfigure}[b]{0.48\textwidth} 
        \centering
        \includegraphics[trim=0mm 0mm 350mm 0mm, clip, width=0.68\textwidth]{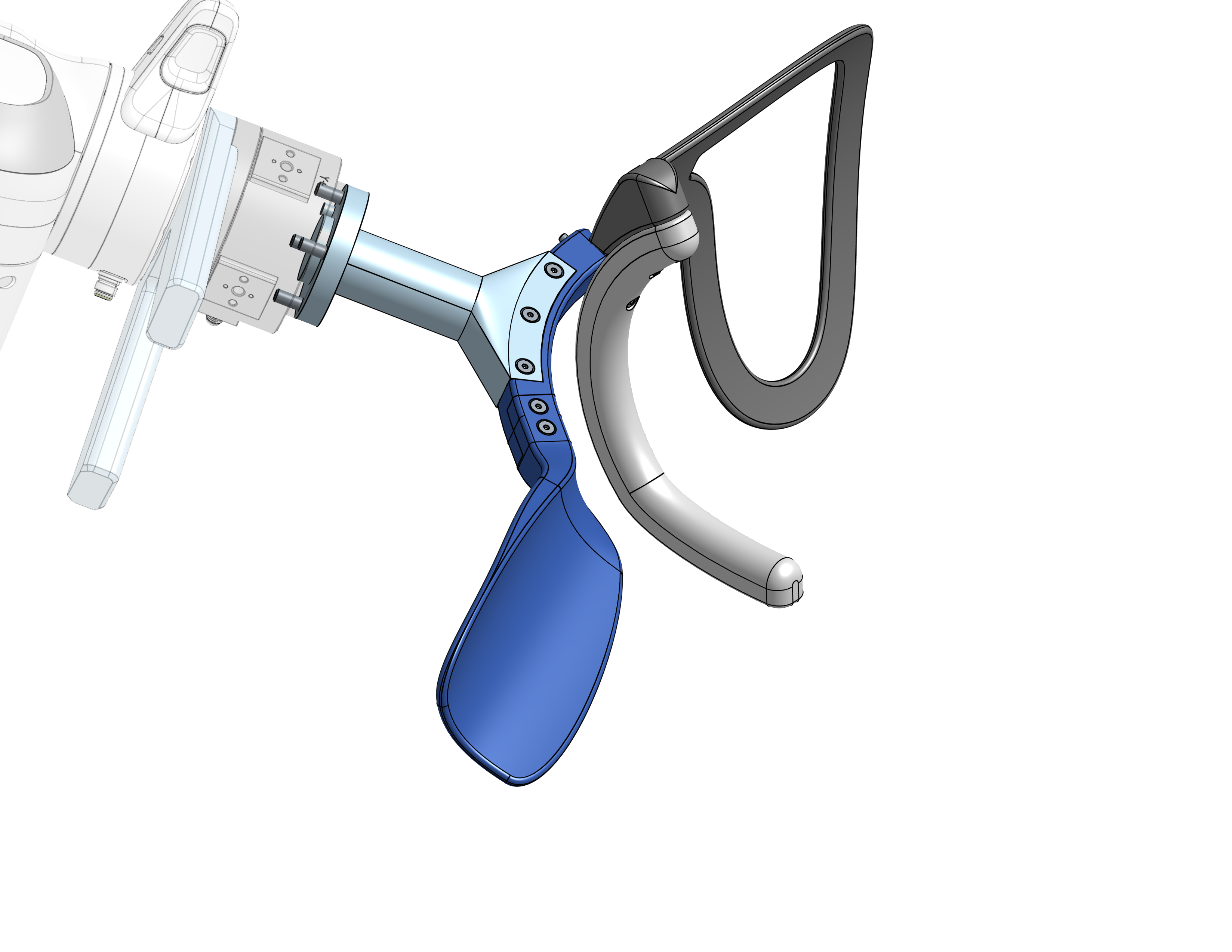}
        \caption{}
        \label{fig:tool_open}
    \end{subfigure}
    \caption{MammoBot end-effectors for patient positioning and support. (\textbf{a}) Right end-effector in the closed configuration (lowered blue ``paddle'') to support the upright, forward-facing posture required for the craniocaudal (CC) view. \textbf{(b)} The same end-effector in the open configuration (raised blue arm-supporting ``paddle'') to assist the angled, oblique stance with the arm raised for the mediolateral oblique (MLO) view.}
    \label{fig:arm_design}
\end{figure*}

\smallskip
\noindent
\textbf{Robotic manipulation platform.}
The manipulation platform comprises two collaborative robotic arms, each equipped with a six-axis force/torque (F/T) sensor mounted at its wrist. These sensors are critical for enabling compliant interaction, providing real-time measurements of the forces and torques exerted while supporting the patient.

\smallskip
\noindent
\textbf{Patient tracking module.} 
For patient tracking, the system employs a thermal perception module. This approach was deliberately chosen over conventional vision-based systems (e.g., RGB cameras) to preserve patient privacy by avoiding the collection of personally identifiable imaging data. The thermal sensors capture the patient's body heat signature, allowing for the non-invasive reconstruction of their posture and position relative to the X-ray mammography equipment and robotic arms.

The system's control loop synergistically integrates feedback from the perception and force-sensing modules:
\begin{itemize}
    \item Posture data from the thermal sensors provides the high-level positional targets for the robotic arms to guide the patient into the correct clinical pose.
    \item F/T sensor readings are fed into a closed-loop controller that modulates the robot's actions in real-time. This allows the system to apply just enough force to support the patient securely without causing discomfort, while also adapting to subtle movements.
\end{itemize}
This dual-feedback mechanism ensures that the physical assistance is both precise and gentle, maintaining the required posture within tight clinical tolerances.

\begin{figure*}[!t]
    \centering
    \begin{subfigure}[b]{0.31\textwidth}
        \centering
        \includegraphics[height=10cm]{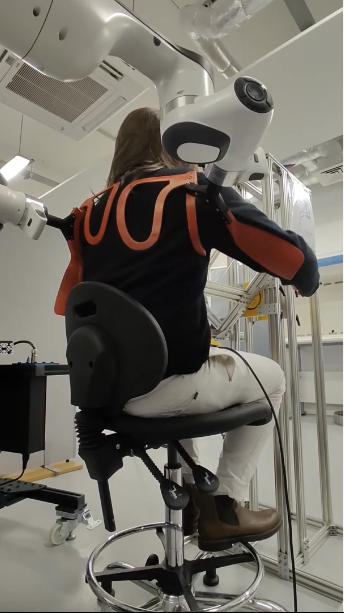}
        \caption{}
        \label{fig:demo_a}
    \end{subfigure}
    \hspace{0.02\textwidth}
    \begin{subfigure}[b]{0.31\textwidth}
        \centering
        \includegraphics[height=10cm]{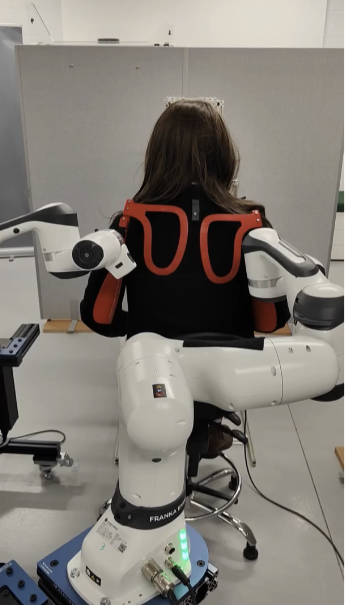}
        \caption{}
        \label{fig:demo_b}
    \end{subfigure}
    \hspace{0.02\textwidth}
    \begin{subfigure}[b]{0.31\textwidth}
        \centering
        \includegraphics[height=10cm]{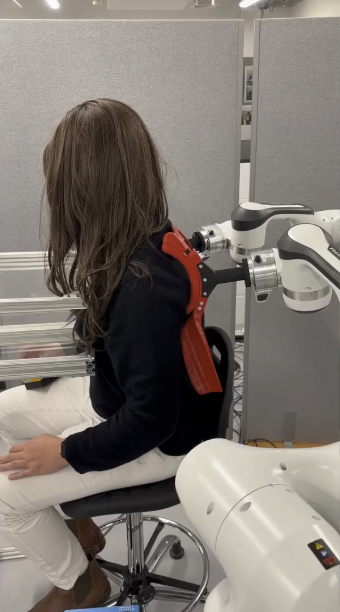}
        \caption{}
        \label{fig:demo_c}
    \end{subfigure}
    \caption{Stills from a demonstration video~\cite{zhu_2026_19348891} showing the proof-of-concept MammoBot system being tested with the help of a research team member in our testbed at the University of York’s Institute for Safe Autonomy. The two 3D-printed end-effectors from Figure~\ref{fig:arm_design} are mounted on Franka Research~3 robotic arms and used to help the researcher adopt the positions required for mammography in front of the X-ray mock-up from Figure~\ref{fig:overall_system}a: (a)~the raised right end-effector paddle brings the person’s arm up for mediolateral oblique (MLO) imaging; (b)~the robot arms and end-effectors support the transition between MLO and craniocaudal (CC) imaging positions; and (c)~the end-effectors push forward slightly to help the person assume the CC-imaging posture.}
    \label{fig:demo_views}
\end{figure*}

\smallskip
\noindent
\textbf{End-effectors.}
A key element of the MammoBot system is a pair of bespoke end-effectors designed for direct patient contact, as shown in Figure~\ref{fig:arm_design}. These components were developed through an iterative, user-centred design process incorporating input from radiographers and users with lived experiences. The resulting end-effector is a multi-contact support tool, ergonomically shaped to comfortably and securely assist patients in maintaining the body positions required for a comprehensive mammographic examination: the upright, forward-facing posture required for \emph{craniocaudal} (CC) imaging (to capture a top-down, ``plan'' view of the breast); and the angled, oblique stances (for the left and right sides) with the arm raised, necessary for \emph{mediolateral oblique} (MLO) imaging (to capture a side-on view that includes the armpit and chest muscle)~\cite{https://doi.org/10.1118/1.4811156}. The physical interaction between the robotic arms and the user during mammography is illustrated in Figure~\ref{fig:demo_views}.

\smallskip
The MammoBot system employs a \emph{layered safety architecture} intended to prevent unsafe operation and enable rapid interruption of robot motion. Safety is ensured through a combination of supervisory control logic, software interlocks, and stop mechanisms operating at multiple levels of authority. At the procedural level, progression through the mammography workflow is governed by explicit gating conditions, ensuring that actions such as robotic motion or X-ray exposure can only occur when prerequisite safety criteria---including verified patient posture, confirmed system readiness, and radiographer approval---are satisfied. At the interaction level, MammoBot supports protective stop functionality through voice commands (e.g., ``stop'') and operator interface controls, allowing either the patient or radiographer to halt robot motion in response to discomfort or unexpected behaviour. These interaction-driven stops are designed to place the system into a \emph{safety-monitored stop state}, where actuator motion is inhibited while maintaining system awareness and enabling safe recovery under operator supervision. Additionally, \emph{hardware-level emergency stops} provide an independent safety layer to immediately disable actuators in response to critical faults or hazardous conditions. While MammoBot is currently a research prototype, its architecture is conceptually aligned with established safety principles for collaborative robots, including protective stops, safety-monitored states, and interlocked operation, as described in standards such as ISO~10218~\cite{ISO10218-1_2011} and ISO/TS~15066~\cite{ISO15066_2016}.

\subsection{Usage scenario}

To illustrate the intended use of the MammoBot system and the interactions between the patient, radiographer, and robotic platform, we describe a possible usage scenario in which the system provides assistance to a wheelchair user named Alex.

\smallskip
\noindent
\textbf{Usage scenario.} 
Alex arrives at the hospital's breast screening department, where she is welcomed and her identity is confirmed. In the screening room, the radiographer explains the mammography process, detailing how the robotic system will support her body to ensure she is comfortable with each step. Once Alex is at ease, a \emph{support team}---comprising the radiographer and, if required, a carer---helps her move into a pre-screening position adjacent to the X-ray machine. While remaining in her wheelchair (which may be a standard hospital model provided for the procedure), she is assisted in assuming a radiographer/carer-supported posture as close as possible to that required for the mammography. 

Next, the radiographer activates the MammoBot system and adjusts the positioning of its two end-effectors under Alex's arms to suit her individual ergonomic requirements (as illustrated on a mannequin in Figure~\ref{fig:system_in_action}). With Alex comfortably positioned, the imaging begins. The end-effectors gently assist her in achieving and maintaining the precise postures required for the three standard imaging views mentioned earlier (the CC view, and both MLO views).

For each view, the radiographer may perform small adjustments via remote control, informing Alex throughout, before triggering the X-ray exposure and verifying the image accuracy. If an image is insufficiently accurate, the process is repeated after further fine-tuning. Between successive views, the radiographer initiates more significant positional adjustments by advancing the robotic workflow, but only after confirming Alex's consent to continue. If at any point Alex or the radiographer deems it necessary, e.g., due to fatigue or discomfort, the session can be abandoned.

Once the imaging is complete or the session is ended, the robotic arms transition to a compliant, low-rigidity state to increase Alex's immediate comfort. This state is maintained for the few seconds until the support team takes over and helps Alex back to her preferred position.

\section{Preliminaries on Hazard Analysis \label{sec:background}}

Hazard analysis techniques are widely used in industrial and safety-critical domains to identify potential risks within complex systems and to derive requirements that mitigate those risks. Traditional approaches to hazard analysis have proven effective in identifying failures related to hardware components, physical processes, and parameter deviations, but they offer more limited support for analysing hazards arising from software behaviour and human interaction~\cite{DAKWAT2018130}. In particular, many established techniques rely heavily on expert brainstorming (e.g., Preliminary Hazard Analysis~\cite{doi:https://doi.org/10.1002/9781118974339.ch7} and Root Cause Analysis~\cite{rooney2004root}) or focus primarily on component-level failures (e.g., Fault Tree Analysis~\cite{5222114} and Failure Mode and Effect Analysis~\cite{stamatis2003failure}), making them less suited to systematically capturing use-related hazards and unsafe interactions in systems with rich user interfaces and human–automation coordination. The rest of this section overviews the two hazard analysis techniques employed by our hazard management methodology.

\subsection{SHARD}

Software Hazard Analysis and Resolution in Design (SHARD)~\cite{521885} is a hazard analysis technique that adapts the core principles of the widely-used Hazard and Operability Study (HAZOP) approach\footnote{HAZOP is a structured, team-based brainstorming technique developed in the 1960s. It uses a set of standard guide words (e.g., ``more'', ``less'', and ``reverse'') to explore deviations from intended design functions and identify potential hazards in complex engineering systems~\cite{crawley2015hazop,kletz2018hazop}.} to address the unique challenges of digital and computer-based systems. SHARD is used to evaluate the intended safety-related behaviour of computer systems that are either safety-critical or safety-related. Although the name emphasises software, SHARD considers the safety aspects of the entire computer-based system rather than focusing solely on the software components.

The primary purpose of SHARD is to support the evaluation of proposed system designs and to help define safety-related requirements that will guide the detailed development phase~\cite{10.1145/381766.381770}. Rather than being a formal safety audit or external assessment, SHARD should be integrated into the design process itself. It is most effective when it is managed and driven by the design team, even if other stakeholders contribute to the analysis. The technique focuses on the flow of information between system components. At a high level, this includes inputs from sensors or external data sources, and outputs to actuators, displays, or other systems. Within the software, SHARD examines how data moves between functions, offering a data-flow perspective that complements the more traditional function-based design view. This shift can help designers and safety engineers uncover potential issues and refine system requirements.

To explore potential hazards, SHARD employs a limited set of guide words, as shown in Table~\ref{tab:guidewords}, that stimulate thinking about deviations from expected information flow behaviours. For each identified deviation, the analysis involves determining its possible causes and assessing whether it could lead to or contribute to a hazardous outcome. If a hazard is identified, the analyst must also evaluate any existing safeguards or mitigation measures present in the design. For any deviation that is both plausible and potentially hazardous, and where existing design measures are insufficient, the analyst should propose suitable actions to enhance the design. 

\begin{table}[!t]
    \centering
    \caption{Original and adapted SHARD guide word definitions for robotics.}
    \sffamily
    \fontsize{7.3}{8.1}\selectfont
    \begin{tabular}{l p{0.73\linewidth}}
    \toprule
         \textbf{Guide Word} & \textbf{Definition} \\ \midrule
         \multirow{2}{*}{Omission} 
         & \textcolor{lightgray}{\textit{Original:} The service is never delivered, i.e., there is no communication.} \\
         & \textit{Adapted:} The robotic service is not performed when required (e.g., the robot fails to detect a user request or does not deliver assistance). \\ \midrule
         
         \multirow{2}{*}{Commission} 
         & \textcolor{lightgray}{\textit{Original:} A service is delivered when not required, i.e., there is an unexpected communication.} \\
         & \textit{Adapted:} A robotic service is performed without a valid trigger (e.g., the robot initiates movement or communication without user command or environmental justification). \\ \midrule
         
         \multirow{2}{*}{Early} 
         & \textcolor{lightgray}{\textit{Original:} The service (communication) occurs earlier than intended. This may be \textit{absolute} (i.e., early compared to a real-time deadline) or \textit{relative} (early with respect to other events or communications in the system).} \\
         & \textit{Adapted:} The robotic service occurs earlier than intended, such as the robot responding before a task condition is met or interrupting the user prematurely. This may be \textit{absolute} or \textit{relative}. \\ \midrule
         
         \multirow{2}{*}{Late} 
         & \textcolor{lightgray}{\textit{Original:} The service (communication) occurs later than intended. As with early, this may be absolute or relative.} \\
         & \textit{Adapted:} The robotic service occurs later than intended (e.g., delayed response to a help request or late delivery of support that affects task performance). \\ \midrule
         
         \multirow{2}{*}{Value} 
         & \textcolor{lightgray}{\textit{Original:} The information (data) delivered has the wrong value.} \\
         & \textit{Adapted:} The information (data) or physical output delivered has the wrong value (e.g., misinterpreted sensor data, incorrect movement parameters or excessive force). \\ 
    \bottomrule
    \end{tabular}
    \label{tab:guidewords}
\end{table}

The analysis process in SHARD is even more structured than in HAZOP, with extra steps to be carried out in the analysis (see~\cite{pumfrey1999principled} for more information). The analysis is recorded in a table with at least the following column headings: \textit{Guide word;} \textit{Deviation}; \textit{Possible Causes}; \textit{Effects}; \textit{Detection and Protection}; \textit{Justification/Design Recommendations}. The guide words are interpreted in terms of a \textit{robotic service}, which may include sensing, decision-making, or actuation steps intended to support a human user. The guide words are used to explore how these intended interactions or communications could fail or deviate from expectations. 

Since SHARD is conducted at the design stage, the quality and reliability of its findings depend heavily on how accurately the final implementation reflects the analysed design. Significant deviations between the implementation and either the original design or the analysts’ understanding of it can greatly undermine the value of the analysis. Although the core concepts of SHARD are straightforward, the effectiveness of the analysis depends on the ability of the analysts to think creatively when applying the guide words and to thoroughly investigate the possible causes and consequences of deviations. This combination of imaginative and systematic thinking is essential for SHARD to meaningfully inform and improve the design process. 

\subsection{STPA \label{subsec:STPA}}

Systems-Theoretic Process Analysis (STPA) is a safety analysis technique grounded in systems theory and modern systems thinking, developed by Levenson~\cite{10.7551/mitpress}. Unlike traditional hazard analysis methods, which were originally developed to predict accidents arising from isolated component failures, STPA recognises that many contemporary safety issues stem instead from unsafe interactions between correctly functioning components~\cite{LEVESON2004237}. These interaction-based accidents may arise from software logic, timing mismatches, incorrect assumptions between system elements, or human–automation coordination problems~\cite{LevesonThomas2018}. For this reason, STPA is often considered a superset of the causal factors identified by classical reliability-based techniques, capable of revealing hazards that would otherwise remain undetected.

In practice, STPA proceeds by modelling the system as a set of interacting controllers, controlled processes, and feedback loops, and by identifying how unsafe control actions may arise within this structure. The analysis examines situations in which control actions are provided when they should not be, omitted when required, provided too early or too late, or applied for an incorrect duration or with inappropriate content. By systematically exploring these deviations in context, STPA enables the identification of hazardous system behaviours and the derivation of safety constraints that guide design and operation.

STPA is typically conducted in three main steps. First, system boundaries are defined and the elements that may influence safety are identified, including both technical components and human operators. Within this system model, potential hazards are examined in terms of unsafe control actions (UCAs), i.e., actions or omissions by system elements that could lead to hazardous states. When analysing use-related risks, a human operator model can be incorporated to represent assumptions about the operator’s knowledge, intentions, and interaction with the system. STPA systematically considers unsafe control actions across four categories: failure to provide a required control action, provision of an unsafe control action, provision of a control action at an incorrect time or sequence, and incorrect duration or persistence of an otherwise safe control action.

In the second step, the analysis focuses on identifying causal factors that could give rise to the identified UCAs. This involves examining the system’s design, interfaces, and operational context to understand why unsafe actions might occur. For human operators, STPA commonly considers deficiencies in feedback (e.g., missing, delayed, or unclear system information), inconsistencies or inaccuracies in the operator’s mental model of system behaviour, and errors or omissions in external information such as procedures, documentation, or communicated instructions.

Finally, STPA supports the derivation of safety requirements and constraints aimed at preventing or mitigating the identified causal factors and unsafe control actions. These requirements translate the analysis into actionable design or operational measures, such as interface improvements, timing constraints, confirmation mechanisms, or procedural safeguards, and can be formulated in a testable manner to support system validation.

\section{Related Work \label{sec:related_work}}

Research on robot safety has traditionally focused on industrial and collaborative robotics, where humans and robots share workspaces but users are typically trained and operate in structured, controlled environments~\cite{7079531,VILLANI2018248}. A substantial body of work has examined how to ensure safe physical interaction through design safeguards (e.g.,~\cite{DBLP:conf/sefm/StefanakosCDAL22,DBLP:journals/scp/GleirscherCDLPA22}), risk assessment methodologies (e.g.,~\cite{ISO12100_2010,ISO13849-1_2015}), and regulatory standards (e.g.,~\cite{ISO10218-1_2011,ISO15066_2016}). Unlike these industrial settings, where safety is framed around collision avoidance and predictable behaviour and all users are trained operators, our work addresses a much more complex socio-technical scenario where the key user is a non-expert, vulnerable patient interacting for the first time with an unfamiliar robotic system. 

As robots have moved beyond constrained industrial settings into collaborative and shared environments, research has increasingly addressed safety in contexts where humans and robots work side by side in closer proximity~\cite{DBLP:journals/arobots/AjoudaniZIAKK18}. This has driven advances in collaborative system design~\cite{DBLP:journals/robotics/MathesonMZFR19}, safety-aware motion planning and control~\cite{DBLP:journals/jim/LiHZP24}, and verification techniques aimed at ensuring reliable operation under uncertainty~\cite{DBLP:journals/csur/LuckcuckFDDF19,DBLP:journals/ijrr/WebsterWADEFP20}. In
contrast to these settings, which often prioritise limiting contact forces, maintaining safe separation distances, or stopping upon contact, our hazard management methodology tackles a scenario where sustained physical contact is a functional requirement, and where the robot must hold and guide the user's upper body into potentially uncomfortable postures.

More recently, attention has shifted toward applications involving vulnerable or non-expert users, such as assistive robotics~\cite{1501143}, rehabilitation systems~\cite{IEC80601-2019}, and healthcare environments where interaction occurs in close physical contact~\cite{DBLP:journals/corr/abs-2209-14041}. In these contexts, users may have limited mobility, reduced strength, cognitive constraints, or heightened anxiety, making safety not only a matter of physical protection but also of usability, comfort, predictability, and trust~\cite{annurev-psych-010416-043958,doi:10.1177/0018720811417254}. This shift has exposed limitations in traditional safety approaches, which often assume users who are attentive, trained, and able to respond quickly to system behaviour. Our research provides a solution that considers these aspects for an assistive robotics application with unique characteristics, as the MammoBot system needs to operate safely in conjunction with both a vulnerable patient (who may be experiencing anxiety or physical discomfort) and a trained radiographer who maintains supervisory control over the mammography procedure.  

A growing body of work (e.g.,~\cite{robotics10020065,10.3389/frobt.2021.667316,GUIOCHET2016225,X-HAZOP}) has explored how safety principles can be adapted for applications involving closer and more direct physical interaction. Cross-domain studies such as~\cite{robotics10020065} highlight the need for updated safety frameworks as robots increasingly operate alongside humans across manufacturing, healthcare, and agriculture. Research focused specifically on assistive robotics has further highlighted the challenges that arise when interaction with non-expert users is continuous and physically sustained. The study in~\cite{10.3389/frobt.2021.667316} examines these issues in the context of robot-assisted dressing and evaluates the applicability of SHARD and STPA for analysing safety in such scenarios. However, even in dressing tasks, the robot does not typically ``hold'' the user in a fixed, forced posture. Our methodology addresses the unique challenge of the trade-off between unavoidable discomfort and diagnosis accuracy: the patient needs to accept a degree of physical compression and sustained holding by the dual-arm robot to ensure a successful diagnostic outcome---a factor not present in daily-living assistance.

Alongside efforts to adapt safety analysis for assistive interaction, researchers have also explored extensions of classical hazard identification methods. The HAZOP-UML approach presented in~\cite{GUIOCHET2016225} demonstrates how deviation-based reasoning can be combined with system modelling techniques to identify operational hazards early in development. By applying structured guidewords to UML representations of robot behaviour, the method helps uncover risks related to unexpected motion, timing errors, and interaction mismatches in environments such as homes and hospitals. More recently, attention has expanded beyond physical safety to include ethical and societal considerations. The X-HAZOP framework~\cite{X-HAZOP} extends HAZOP-style reasoning by introducing participatory, scenario-based techniques to identify ethical hazards in assistive robotics, including concerns related to autonomy, dignity, bias, and marginalisation. Our methodology builds upon these structured approaches but is unique in its integrated pipeline: we begin with process co-design involving stakeholders, followed by a dual-technique hazard analysis using SHARD and STPA, and conclude with process tuning specifically informed by the outcomes of that analysis. This ensures that safety is not just an ``add-on'' but a design driver that accounts for the fact that a patient will encounter an unfamiliar robot only once every three years.

At the same time, system-theoretic approaches like STPA have been applied to complex software-intensive systems~\cite{OGINNI2023106275,10173978}, including safety-critical medical systems~\cite{SEFM-hazard-analysis}. Together, these studies underline the suitability of system-theoretic and interaction-focused safety methods for analysing socio-technical systems where safety depends on coordination between humans, software, and automated components. Our methodology leverages these insights and demonstrates how SHARD and STPA can be combined to capture hazards that arise from the intersection of robotic control logic, sensor-driven physical manipulation, and the human-in-the-loop coordination between the radiographer and the patient.

In parallel, standards and regulatory frameworks continue to shape safety practices across domains. For example, ISO/TS 15066 provides guidance on collaborative robot safety through the definition of safeguards such as force limits, speed monitoring, and separation distances, and highlights the importance of aligning hazard analysis outcomes with design decisions~\cite{CHEMWENO2020104832}. While these standards provide a foundation, they lack the specificity required for the close interaction present in our application domain. This gap motivates the methodology proposed in our work, which moves beyond general force-limiting to a context-aware safety architecture tuned for the specific clinical workflow associated with mammography.

\section{Methodology \label{sec:methodology}}

The methodology adopted in this work is grounded in a process-driven and multidisciplinary approach. First, a realistic representation of the mammography workflow was constructed through collaboration with various stakeholders to capture the key human–robot interactions involved in the breast screening procedure. Based on this process model, an initial set of system requirements was derived to describe the intended functionality and constraints of the MammoBot system. Second, SHARD was applied to systematically explore technical and procedural deviations across the workflow. Third, STPA was used to identify unsafe control actions arising from human–system interaction. Finally, the findings from both analyses were brought together to inform the subsequent refinement of the system behaviour and requirements.

\subsection{Process design \label{subsec:process_design}}

The process was developed collaboratively with breast-screening radiographers, radiologists, roboticists and users with lived experiences, to ensure that it accurately captures the practical, clinical, and human considerations involved in the use of the MammoBot system. The design began with a series of workshops and discussions involving a radiographer from our team, whose clinical expertise was essential to map the established mammography workflow. This step helped to identify key stages of the screening process, typical decision points, and critical safety and comfort considerations for patients.

To ensure the process realistically reflected the technical capabilities and limitations of the robotic system, we consulted with roboticists involved in the development of MammoBot. Their input helped define the boundaries of automation, identify which tasks could be safely delegated to the robotic arms, and specify where operator supervision or confirmation would remain essential. These discussions guided the inclusion of control and validation steps such as posture detection, trajectory planning, and motion interlocks.

In parallel, the process design incorporated feedback from meetings held with representatives from the Spinal Injuries Association, one of the UK's leading spinal cord injury charities, to ensure accessibility and inclusivity for users with limited mobility. Through these consultations, we learned about the physical and psychological challenges that individuals with spinal injuries or other mobility impairments may face when positioning for mammography. These insights informed the design of patient preparation and positioning steps, the inclusion of comfort checks, and the emphasis on human–robot interaction points where patients can communicate needs or discomfort.

By combining clinical knowledge, robotic expertise, and lived experience, the resulting process model represents a realistic and patient-centred workflow. The process was formalised using a UML activity diagram~\cite{OMG_UML}, a widely used structured notation for representing sequences of actions, decision points, and human–system interactions~\cite{rumpe2016modeling}. Such representations are commonly used in safety-oriented design to make control flow and responsibilities explicit, and to support systematic hazard analysis~\cite{GUIOCHET2016225}. In this work, the model serves as the foundation for the subsequent hazard analyses, enabling the systematic identification of potential deviations and unsafe interactions---and their associated causes and mitigations---across both the robotic and human-in-the-loop aspects of the procedure.

\begin{figure}
    \centering
    \includegraphics[width=0.955\linewidth]{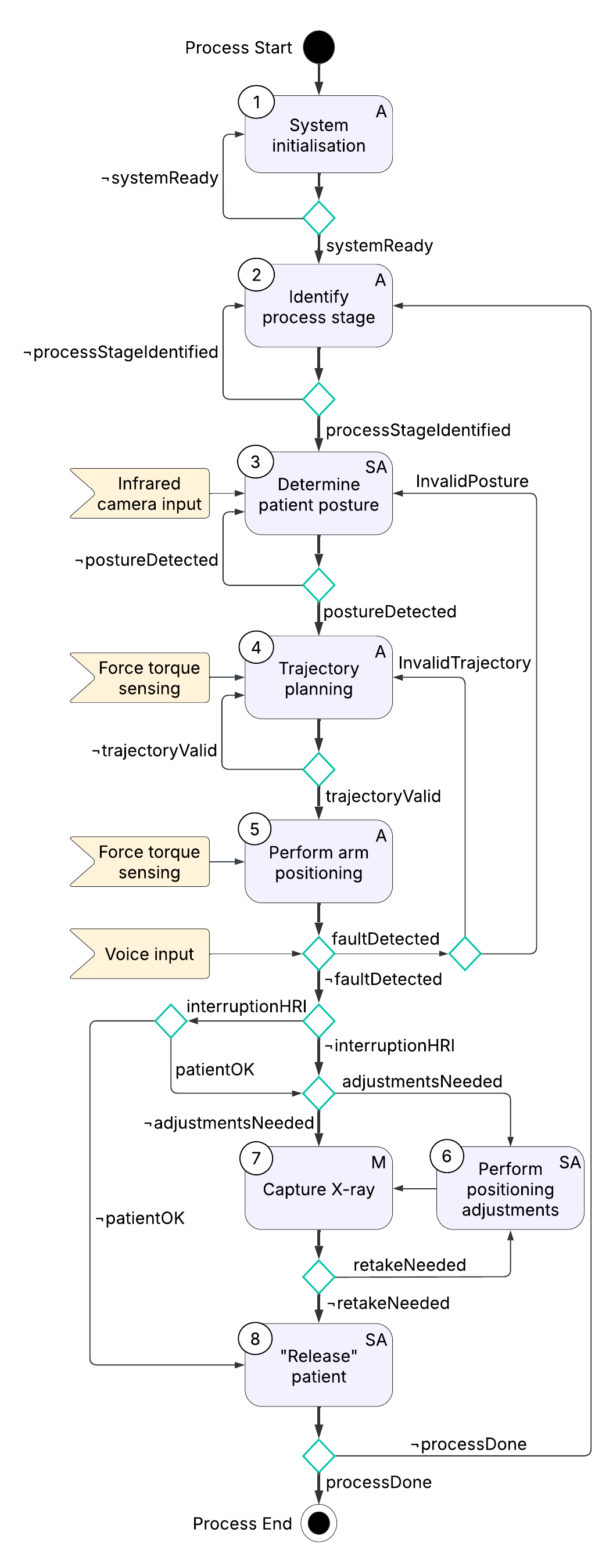}
    \caption{UML activity diagram of the MammoBot process. Rounded rectangles denote actions, diamonds represent decision points, with negated guards (e.g., $\neg\:\! \mathsf{patientReady}$) indicating false outcomes. Sensor inputs and human commands are shown explicitly to highlight human–robot interaction. Actor annotations (A: automated; M: manual; SA: semi-automated) indicate control responsibility at each stage.}
    \label{fig:process_diagram}
\end{figure}

The process developed for the MammoBot system, depicted in Figure~\ref{fig:process_diagram}, outlines the main operational phases of a patient imaging session. Each phase represents a key interaction between the robot, the radiographer, and the patient, designed to maintain safety, efficiency, and comfort. The workflow integrates automated functions with human supervision and consent at each critical stage. Below is a description of this MammoBot process.

\smallskip
\noindent
\textbf{1) System initialisation:} This step represents the start-up and readiness phase of the MammoBot system. When powered on, the robot performs a sequence of internal checks to ensure that hardware, sensors, and software modules are functioning correctly. Configuration files are loaded, safety systems are tested, and communication with the operator interface is verified. Only once all subsystems report a safe and stable status does the system become ready for use. This step ensures that subsequent operations begin from a known, safe baseline. As shown in the activity diagram, progression beyond this stage is explicitly gated by a readiness check; if the guard \textsf{systemReady} does not hold, the workflow remains in this state and prevents further actions from being initiated.

\smallskip
\noindent
\textbf{2) Identify process stage:} In this step, the system identifies which stage of the mammography workflow is currently active (i.e., craniocaudal, left-view mediolateral oblique, or right-view mediolateral oblique imaging). This helps both the system and the radiographer stay synchronised in the process. The system uses contextual information such as image sequence number, arm position, and operator inputs to determine the current stage. This prevents confusion between stages, ensuring the system does not begin repositioning to an already captured X-ray imaging position. If the process stage cannot be confidently identified (i.e., the guard \textsf{processStageIdentified} does not hold), a decision point in the workflow prevents progression and requires clarification or confirmation before continuing.

\smallskip
\noindent
\textbf{3) Determine patient posture:} This step involves the acquisition and interpretation of patient posture data using sensors such as infrared cameras. The system analyses these data to understand the patient’s orientation and proximity relative to the imaging equipment. The radiographer oversees this process to confirm that the detected posture matches clinical requirements. Accurate posture detection is crucial to guide the systems’s subsequent movements safely and to reduce the need for manual repositioning, particularly for patients with limited mobility. Posture detection is subject to validation; if posture cannot be detected or is deemed invalid (i.e., the guard \textsf{postureDetected} is false), the workflow loops within this step rather than advancing prematurely.

\smallskip
\noindent
\textbf{4) Trajectory planning:} Here, the system computes the path that the robotic arms should follow to achieve the desired positioning for imaging. This involves translating the patient’s detected posture into precise, collision-free joint motions that respect both comfort and safety constraints. Trajectory planning incorporates physical limits of the robot arms, proximity sensors, and safety envelopes around the patient. The plan is reviewed and approved by the operator before execution, to ensure the movements remain clinically appropriate and patient-safe. As reflected in the diagram, the planned trajectory is explicitly checked for validity; only trajectories for which the guard \textsf{trajectoryValid} holds are allowed to progress to execution.

\smallskip
\noindent
\textbf{5) Perform arm positioning:} In this step, the system executes the planned motion to assist the patient into the correct posture for imaging. The robot arms move slowly and compliantly to support or adjust the patient’s position, with the radiographer supervising the action through the human–robot interface. Safety mechanisms continuously monitor contact forces and positions to prevent excessive pressure or overextension. This is critical to balance robotic precision with patient comfort and reassurance. If a fault is detected (i.e., \textsf{faultDetected} holds) or an interruption is requested (i.e., \textsf{interruptionHRI} holds), motion is halted and the workflow transitions into a safe recovery path rather than continuing automatically.

\smallskip
\noindent
\textbf{6) Perform positioning adjustments:} After the initial positioning, fine adjustments may be needed to optimise the imaging geometry (i.e., the guard \textsf{adjustmentsNeeded} is true). These small corrections compensate for patient movement, minor posture deviations, or variations in anatomy. The system performs these adjustments under operator command or based on automatic image quality feedback. Keeping adjustments smooth and minimal reduces patient discomfort while ensuring that the imaging region is correctly aligned for accurate diagnostic results. 

\smallskip
\noindent
\textbf{7) Capture X-ray:} At this stage, the radiographer initiates the X-ray exposure once the patient and robot arms are both stable. The robot arms hold position to maintain consistent geometry during the short exposure period. System interlocks ensure that exposure cannot be triggered until all safety and readiness conditions are satisfied (e.g., patient consent, proper alignment, no movement). This step produces the diagnostic image used for assessment and is repeated as needed for different views. Following exposure, a decision point evaluates whether the image quality is sufficient; if the guard \textsf{retakeNeeded} holds, the workflow returns to the adjustment and positioning stages before attempting another capture.

\smallskip
\noindent
\textbf{8) ``Release'' patient:} This final stage marks the end of the imaging phase. The robot transitions into a compliant, relaxed state, allowing the patient to move freely and comfortably. Release is conditioned on the completion of required checks to ensure that no further adjustments or retakes are pending.

\smallskip
\noindent
The process is repeated until all required X-ray images have been successfully acquired, or until the session is terminated early due to clinical or patient-related factors, such as pain, discomfort, or fatigue. In the activity diagram, this is represented by the guard \textsf{processDone}; only when this condition holds does the workflow terminate.

\subsection{System requirements}

Following the process design activities, a set of high-level system requirements was defined to capture the intended functions, safety behaviours, and interaction principles of the MammoBot system. These requirements were derived from the collaboratively developed process model and reflect clinical practice, robotic system capabilities, and accessibility considerations. They established a baseline specification for the system prior to safety analysis.

The requirements are organised into three complementary categories: functional requirements, which describe the core sensing, positioning, control, and data-management capabilities needed to support the breast-screening workflow; safety requirements, which define constraints and mechanisms intended to prevent harm and ensure safe operation under faults, uncertainty, or unexpected events; and human--robot interaction (HRI) requirements, which address communication, feedback, remote control, and patient-centred interaction.

Table~\ref{tab:system_requirements} summarises the key system requirements identified at this stage. These requirements informed the subsequent hazard analyses, providing a reference framework against which deviations from intended behaviour could be systematically explored.

\begin{table}[!t]
    \centering
    \caption{MammoBot system requirements.}
    \fontsize{7.3}{8.1}\selectfont
    \rowcolors{2}{gray!10}{white}
    \sffamily
    \begin{tabular}{l p{7.5cm}}
        \toprule
        \textbf{ID} & \textbf{Description} \\ 
        \hiderowcolors
        \midrule
        \multicolumn{2}{c}{\textbf{Functional requirements}}
        \\ \midrule
        \showrowcolors
        $R_{1}$  & The system shall respond to radiographer commands (manual or voice input) in <1 second \\  
        $R_{2}$  & The system shall be able to determine the stage of the screening process (positioning, imaging, transitioning, etc.) \\ 
        $R_{3}$  & The system shall be able to identify and adjust the position of the end-effectors under the patient's arms based on input from the radiographer or sensors \\ 
        $R_{4}$  & The system shall assist the patient in achieving and maintaining the required posture with an accuracy of ±5 mm \\  
        $R_{5}$  & The system shall enable fine adjustments to end-effector positioning (±2 mm) as instructed by the radiographer \\ 
        $R_{6}$  & The system shall detect major posture changes and assist in repositioning between successive X-rays \\  
        $R_{7}$  & The system shall detect the patient’s body geometry (arms, torso, etc.) and adapt to their unique needs \\  
        $R_{8}$  & The system shall automatically log session data (e.g., posture adjustments) for quality assurance \\  
        $R_{9}$  & The system shall maintain a patient profile with preferences for positioning or prior session data to improve future sessions \\  
        \hiderowcolors
        \midrule
        \noalign{\global\rownum=2}
        \multicolumn{2}{c}{\textbf{Safety requirements}}
        \\ \midrule
        \showrowcolors
        $R_{10}$  & The system shall detect the patient's fatigue or discomfort during the procedure using pressure or motion sensors \\
        $R_{11}$  & The system shall provide gentle, smooth movement of the robotic arms to ensure patient comfort \\
        $R_{12}$  & The system shall reduce rigidity to provide patient comfort after imaging is complete or in between steps of the process \\  
        $R_{13}$  & The system shall identify and report procedural errors (e.g., incorrect positioning) to the radiographer \\  
        $R_{14}$  & The system shall allow the session to be paused or stopped immediately upon request from the radiographer or patient \\  
        $R_{15}$  & The system shall employ a safety executive function to avoid harm to the patient, ensuring compliance with medical device regulations \\  
        $R_{16}$  & The system shall ensure imaging accuracy by maintaining the required posture during X-rays and notifying the radiographer of deviations \\ 
        \midrule
        \multicolumn{2}{c}{\textbf{HRI requirements}}
        \\ \midrule
        $R_{17}$  & The system shall allow radiographers to remotely control the positioning of the end-effectors without physical intervention \\
        $R_{18}$  & The system shall provide feedback to the radiographer and patient during adjustments (e.g., auditory) \\
        $R_{19}$  & The system shall initiate patient-centric HRI strategies (e.g., reassuring messages or guidance) to keep the patient at ease \\ 
          
        \bottomrule
    \end{tabular}
    \label{tab:system_requirements}
\end{table} 

\subsection{Applying SHARD to the MammoBot process \label{subsec:SHARD}}

To evaluate the safety and reliability of the MammoBot process, the SHARD methodology was applied to the activity diagram developed during process design. The diagram served as the foundation for the hazard analysis, with each node representing a discrete action or decision. For every node, the five SHARD guidewords (Omission, Commission, Early, Late, and Value) were applied to explore possible deviations from the intended behaviour. Instances where some of the guidewords were not applicable were explicitly noted to maintain analytical transparency.

The SHARD methodology was applied collaboratively. The analysis combined inputs from clinical experts (radiographers), who clarified the expected flow and critical safety checks in existing mammography practice; roboticists, who defined the operational capabilities, limitations, and control logic of the robotic system; computer scientists, who contributed expertise in system integration, data handling, and software reliability; and legal experts, who provided guidance on ethical and regulatory considerations surrounding patient data, safety standards, and accountability in autonomous and semi-autonomous medical systems. This multidisciplinary approach ensured that the identified hazards and mitigations addressed both technical and human factors while remaining compliant with legal and ethical frameworks for clinical deployment.

For each deviation, the analysis identified the possible causes, potential effects, detection or protection mechanisms, and design recommendations. These findings were consolidated into Table~\ref{tab:SHARD_table} in~\ref{app1}, which highlights where risks are primarily technical (e.g., sensor or communication faults) or operational (e.g., delayed confirmations, or incorrect inputs). Each entry also includes a qualitative hazard level to support prioritisation during design refinement. The hazard levels indicate the likely severity and immediacy of the consequences associated with each deviation. ``High'' hazards correspond to situations with a direct or credible path to patient harm, unsafe motion, unintended exposure, or operation outside defined safety limits. ``Medium’’ hazards represent conditions that may not be immediately dangerous but could propagate into unsafe states if left unaddressed. ``Low'' hazards denote contained safety-relevant issues that mainly lead to degraded operation or workflow interruption. Finally, the ``Annoyance’’ level captures non-physical impacts such as delays, frustration, workflow inefficiencies, or mild patient anxiety. This classification helps distinguish between deviations that directly threaten safety and those that primarily affect comfort, usability, or procedural efficiency, while still acknowledging their importance in patient-centred clinical environments.

This qualitative, ordinal classification supports structured prioritisation during design refinement and is consistent with severity-based reasoning approaches used in other safety-critical domains, where harms are commonly organised into ordered levels rather than quantified precisely. For example, injury-based scales such as the Abbreviated Injury Scale, widely used in medical and automotive safety research, classify outcomes according to their seriousness to support systematic risk evaluation~\cite{Loftis28122018}. Although the hazard levels used in this work are not intended to map directly to clinical injury categories, they provide an analogous conceptual structure for reasoning about the relative seriousness of potential consequences in a human–robot interaction context.

While SHARD prescribes a fixed set of guidewords, their applicability depends on the semantic role of each node in the process. In this analysis, nodes were interpreted according to whether they represent physical actions, human-in-the-loop checks, automated evaluations, or logical completion conditions. For action-oriented nodes, all five guidewords were applicable, as deviations may occur in execution, timing, or parameter values. For decision and evaluation nodes, however, some guidewords collapse conceptually; for example, ``Early'' and ``Late'' do not represent distinct hazards for purely logical decisions, where delayed or premature evaluation manifests instead as an incorrect or missing decision. In such cases, deviations are more appropriately captured using Omission, Commission, or Value. These distinctions were applied consistently to ensure that the analysis focused on meaningful hazard modes rather than mechanically applying guidewords where they do not generate new insight.

The SHARD analysis revealed several recurring vulnerability patterns across the MammoBot process. Timing-related deviations were particularly prominent in stages involving close coordination between the radiographer, patient, and robotic system, such as posture determination, arm positioning, and X-ray capture. These findings emphasise the importance of explicit gating conditions and stable-state verification before allowing progression to safety-critical actions. Value-related deviations frequently arose from configuration, calibration, or threshold errors, highlighting the need for robust parameter validation, clear system feedback, and conservative default settings.

In addition to these recurring deviation types, the analysis showed that hazards are not evenly distributed across the workflow. Early-stage nodes, such as system initialisation, are dominated by technical and configuration-related risks, whereas mid- and late-stage nodes increasingly involve operational and state-management deviations, including timing mismatches, incorrect sequencing, or inconsistencies between system state and process context. This progression underscores the need for safety mechanisms that evolve with the task context, shifting from configuration assurance and system readiness checks in the early phases to stronger control-flow validation and state consistency safeguards as the procedure advances.

Beyond identifying individual hazards, the SHARD analysis directly informed the refinement of system requirements, as discussed in more detail in Section~\ref{subsec:Refinement}. Several existing requirements were strengthened by introducing explicit timing constraints, confirmation dependencies, and non-bypassable safety checks, while additional requirements were derived to address gaps revealed by the analysis. These include constraints on when motion or exposure may be enabled, requirements for stable posture verification, and mechanisms for preventing premature progression through the workflow. By systematically linking deviations to design recommendations, SHARD provided traceability between the process model, identified hazards, and the evolving system specification.

\subsection{User Error Analysis using STPA}

\begin{figure*}[!t]
    \centering
    \includegraphics[trim=6mm 6mm 6mm 6mm, clip, width=\linewidth]{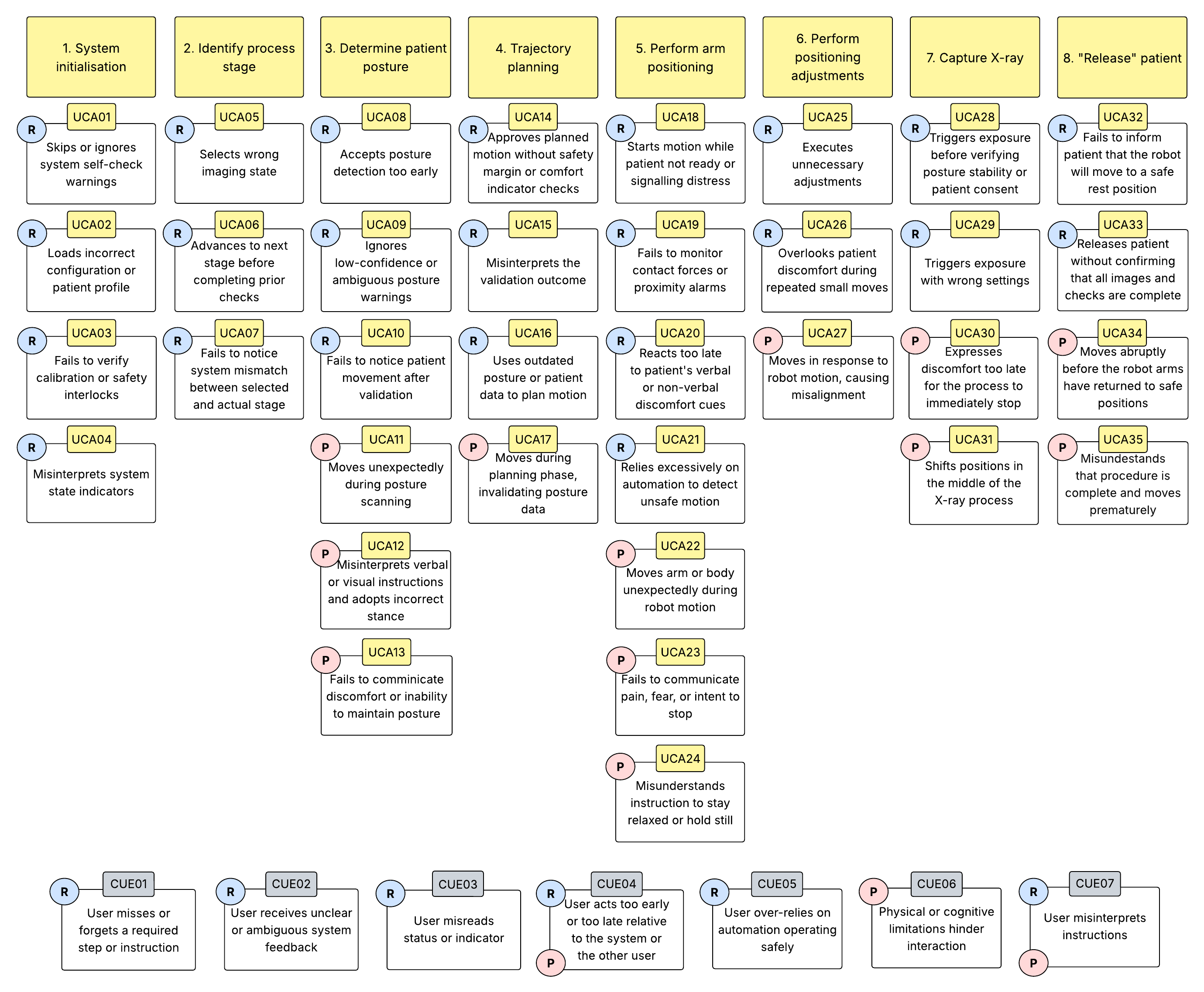}

    \vspace*{2mm}
    \caption{STPA-based user error analysis for the MammoBot breast screening process. For each action node, potential UCAs are identified and listed beneath the corresponding stage. Each UCA is annotated with a circular role marker indicating the primary source of the unsafe action: \textbf{R} denotes a radiographer-related user error, while \textbf{P} denotes a patient-related user error. The bottom row of the diagram lists CUEs, which are generic user-error patterns that may occur in any step of the process, regardless of the specific task.}
    \label{fig:STPA}
\end{figure*}

To strengthen the analysis of the MammoBot process further, we employed STPA as a safety analysis technique complementary to SHARD. STPA is particularly suited to this context due to the close coupling between human users and robotic assistance, the reliance on software-driven control logic, and the need to coordinate patient actions, clinician decisions, and automated system behaviour within a clinical workflow.

MammoBot is an example of a socio-technical system in which radiographers, patients, robotic subsystems, sensors, and clinical workflows interact closely. As noted in the research literature, applying STPA to such systems requires adaptation to the specific operational context and task structure, because the range of possible human actions is finite but large and highly dependent on workflow, user capability, and interface design~\cite{BOLBOT2019179}. Therefore, rather than analysing human behaviour in the abstract, we grounded our STPA in the concrete task scenario defined in the process diagram. This scenario provides a clear view of the essential actions humans must perform or supervise during the procedure. STPA enables an initial abstraction of the types of unsafe user behaviours that could lead to hazards. These may include issuing a control action at the wrong time, failing to issue a required action, misinterpreting system state, or being unable to provide critical input due to physical or cognitive constraints. The analysis therefore captures unsafe control actions arising from both radiographer interactions and patient behaviour, as well as more general user-error patterns that can occur across multiple tasks.

The diagram from Figure~\ref{fig:STPA} shows the output of STPA carried out for the MammoBot breast-screening workflow, focusing on the eight major nodes of the process introduced in Section~\ref{subsec:process_design}. Under each node, the diagram lists the specific unsafe control actions (UCAs) that could occur at that node. In this work, STPA is applied to unsafe control actions issued by human users (radiographers and patients) while interacting with the MammoBot system. These human-issued UCAs represent unsafe actions or omissions that may lead to hazardous system states during breast screening. System-level and technical faults identified earlier using SHARD (see Section~\ref{subsec:SHARD}) are therefore not repeated here, allowing the STPA to complement the preceding analysis by concentrating specifically on human–system interaction risks.

Each UCA is labelled with R (radiographer error) or P (patient error) to distinguish the two user roles and reflect their different interactions with the system. The UCAs(R) identified include skipping self-check warnings during system initialisation, accepting posture before stabilisation, approving a motion plan without reviewing safety margins, failing to monitor patient discomfort, or triggering an X-ray before verifying posture stability, among others. UCAs(P) include moving unexpectedly during scanning or positioning, misinterpreting instructions, failing to communicate discomfort, or moving prematurely before the robot has returned to a safe rest position. 

Along the bottom of the diagram, the seven common user errors (CUE01–CUE07) capture recurring unsafe user behaviours that may arise at multiple stages of the process. These include missing or forgetting a required step (CUE01), receiving unclear or ambiguous feedback from the system (CUE02), misreading system status indicators (CUE03), acting too early or too late relative to system state (CUE04), over-trusting automation (CUE05), physical or cognitive limitations affecting interaction (CUE06), and misinterpreting system or user instructions (CUE07). Unlike the UCAs, which are associated with specific process nodes, CUEs are cross-cutting and may occur at different points in the workflow. A complete list of identified UCAs/CUEs, together with their causes, effects, detection mechanisms, and design recommendations, is provided in Table~\ref{tab:UCA_CUE_table} from~\ref{app2}.

The STPA analysis provides several important insights. First, by mapping each task to its unsafe actions, it reveals where the workflow is most error-prone. Tasks such as \textit{determine patient posture} and \textit{performing arm positioning} contain dense clusters of both radiographer and patient errors, whereas earlier tasks like system initialisation show far fewer. This helps identify the stages that require stronger feedback, clearer interaction cues, or additional safety checks. 

A further insight is gained from distinguishing radiographer and patient errors rather than treating ``the user'' as a single entity, showing that radiographer errors tend to be related to timing, interpretation, attention, or over-reliance on automation, while patient errors often stem from movement, misunderstanding, discomfort, or limited ability to communicate. This separation makes it possible to design role-specific mitigations, such as interface redesign for the radiographer or improved communication and comfort support for the patient.

This analysis also exposes critical coordination challenges between automation and human supervision. Several unsafe control actions arise at transition points where the robotic system assumes conditions such as stable posture, immobility, or confirmed readiness, while users may act under uncertainty, fatigue, or incomplete feedback. These mismatches help explain why unsafe actions can occur even when individual system components operate as intended, and they emphasise the need for design measures that reduce reliance on precise human timing or interpretation alone.

Importantly, the STPA analysis provides a direct basis for defining and justifying safety constraints. By linking unsafe actions to specific process nodes and recurring user errors, the analysis supports constraints such as blocking motion or exposure until required confirmations are complete, enforcing stabilisation periods following posture changes or interruptions, and automatically pausing or slowing the robot in response to unexpected patient movement. These constraints are traceable to identified risks and directly inform the solution refinement described in the following section.

\subsection{Solution refinement \label{subsec:Refinement}}

The safety analyses carried out using SHARD and STPA were used to refine the MammoBot process and system design beyond the initial functional specification. The refinement process systematically translated identified deviations, unsafe control actions, and human-factor patterns into concrete design improvements, safety constraints, and requirement clarifications.

\subsubsection{Refinements from SHARD analysis}

\vspace*{3mm}
The SHARD analysis revealed several recurring categories of deviation across the process, which informed system-level refinements.

\paragraph{Strengthening preconditions and gating logic} Many high-risk deviations arose from actions occurring when prerequisite conditions were not yet satisfied (e.g., motion starting before posture validation, exposure before stability confirmation). As a result, key transitions in the process were refined to include explicit gating conditions that must be satisfied before progression. These include calibration completion, patient readiness confirmation, posture validity, and trajectory validation. Making these preconditions explicit reduces reliance on operator vigilance alone and prevents unsafe sequencing.

\paragraph{Improving fault detection, timing control, and recovery} SHARD identified risks associated with delayed, missed, or spurious fault detection, particularly during arm motion and exposure-related steps. This led to refinements that prioritise fault monitoring as a mandatory, non-bypassable function, with bounded detection times and clear recovery guidance. Timeouts, watchdogs, and progress feedback were emphasised to avoid silent failures and prolonged patient discomfort.

\paragraph{Ensuring robustness against incorrect or stale values} Value-related deviations, such as incorrect posture parameters, trajectory scales, or exposure settings, were consistently associated with high hazard levels. To mitigate these risks, the refined design emphasises plausibility checks, strong typing of contextual data, and explicit operator previews before execution. This ensures that incorrect internal values are detected early and cannot propagate silently into hazardous actions.

\subsubsection{Refinements from STPA analysis}

\vspace*{3mm}
While SHARD exposed deviations in process execution, STPA revealed how predictable patterns of human interaction could lead to unsafe system states.

\paragraph{Role-specific safeguards for radiographers and patients} By distinguishing radiographer-issued and patient-issued unsafe control actions, the refinement process supports role-appropriate mitigations. Radiographer-facing refinements focus on clear system feedback, staged confirmations, and prevention of premature approvals, particularly for motion and exposure. Patient-facing refinements prioritise comfort, communication, and low-effort interruption mechanisms, recognising physical limitations and anxiety during screening.

\paragraph{Mitigating common user-error patterns} The common user error categories (CUE01–CUE07) highlight recurring issues such as missed steps, misinterpretation of feedback, acting too early or too late, and over-reliance on automation. Rather than addressing these through training alone, the refined design embeds countermeasures directly into the system, such as enforced sequencing, stabilisation windows, explicit status indicators, and automation transparency.

\paragraph{Coordinating automation and human supervision} The analysis exposed mismatches between what the robotic system assumes (e.g., stillness, readiness, confirmation) and what humans realistically do (e.g., move slightly, hesitate, or misinterpret cues). Refinements therefore focus on tighter coordination between automation and human oversight, including slowing or pausing automation when uncertainty is detected, and requiring human confirmation at safety-critical points.

\subsubsection{Impact on system requirements}

\vspace*{2mm}
The refinements derived from SHARD and STPA directly informed updates to the system requirements defined earlier in the design process. Existing requirements were clarified with additional constraints on timing, validation, and interaction, while new safety-oriented requirements were introduced to enforce non-bypassable checks, stabilisation periods, and role-specific confirmations. This ensures that all refinements are traceable to specific system requirements rather than remaining as informal recommendations.

Table~\ref{tab:refined_requirements} summarises how these refinements are reflected at the level of system requirements. Each row retains the original requirement identifier, preserving continuity with the initial specification, while the refined description makes explicit the additional constraints introduced by the safety analyses. The methodology column indicates whether a refinement was motivated primarily by SHARD, STPA, or both, showing how technical deviations and human-interaction risks contributed to each requirement update.

\begin{table*}[!t]
\centering
\fontsize{7.3}{8.1}\selectfont
\sffamily
\caption{Refinement of MammoBot system requirements based on SHARD and STPA analyses. Each requirement retains its original identifier, while the methodology column provides direct traceability to the safety analysis (SHARD and/or STPA) that informed the refinement.}
\rowcolors{2}{gray!10}{white}
\begin{tabular}{p{0.5cm} p{5.5cm} p{8.2cm} p{2.2cm}}
\toprule
\textbf{ID} & \textbf{Original requirement summary} & \textbf{Refined requirement / added constraint} & \textbf{Methodology} \\
\midrule
$R_{1}$ & Respond to radiographer commands in <1 second &
Command execution shall only occur once all safety preconditions are satisfied (initialisation complete, posture valid, patient ready). The response-time requirement applies after gating. &
SHARD \\
$R_{2}$ & Determine the stage of the screening process &
Stage identification shall be validated against contextual cues (arm pose, image index, patient readiness). Inconsistencies shall trigger operator confirmation rather than automatic progression. &
SHARD \\
$R_{3}$ & Identify and adjust end-effector position based on radiographer or sensors &
End-effector adjustments shall be inhibited when posture confidence is low or an interruption is active. Manual overrides shall not bypass safety constraints. &
SHARD, STPA \\
$R_{4}$ & Assist patient posture with ±5 mm accuracy &
Posture assistance shall only be applied after posture stability is confirmed for a minimum dwell time. Detected instability shall pause motion automatically. &
SHARD, STPA \\
$R_{5}$ & Enable fine adjustments (±2 mm) &
Fine adjustments shall be previewed before execution and constrained within predefined safety envelopes to prevent excessive or unintended motion. &
SHARD \\
$R_{6}$ & Detect major posture changes between X-rays &
Major posture changes shall automatically pause the workflow and require revalidation before progression to planning or exposure. &
SHARD, STPA \\
$R_{7}$ & Detect patient body geometry and adapt accordingly &
Body geometry estimation shall be revalidated following interruptions or significant patient movement before further motion is permitted. &
SHARD \\
$R_{8}$ & Automatically log session data &
Session logs shall include timestamps for posture changes, interruptions, faults, and confirmations to support traceability and auditability. &
SHARD \\
$R_{9}$ & Maintain patient profile and preferences &
Patient preferences shall inform comfort-related behaviour but shall not override safety constraints or validation checks. &
SHARD \\
$R_{10}$ & Detect patient fatigue or discomfort &
Patient wellbeing shall be explicitly checked before each exposure and major repositioning step, requiring radiographer confirmation and patient assent where possible. &
STPA \\
$R_{11}$ & Provide gentle, smooth robotic arm movement &
Motion speed and force limits shall adapt dynamically based on detected patient movement, interruption requests, or uncertainty in posture estimation. &
SHARD, STPA \\
$R_{12}$ & Reduce rigidity after imaging or between steps &
Rigidity reduction shall be triggered automatically after exposure completion or upon detection of patient discomfort. &
STPA \\
$R_{13}$ & Identify and report procedural errors &
Procedural errors shall be reported with clear recovery guidance and shall block further progression until acknowledged or resolved. &
SHARD \\
$R_{14}$ & Allow immediate pause or stop &
Pause or stop requests shall be non-bypassable, system-wide, and handled through a high-priority safety channel independent of UI state. &
SHARD \\
$R_{15}$ & Employ a safety executive function &
The safety executive shall enforce non-bypassable validation of posture, trajectory, readiness, and exposure conditions prior to motion or imaging. &
SHARD \\
$R_{16}$ & Ensure posture accuracy during X-rays &
X-ray exposure shall only be enabled when posture stability, arm immobility, and patient readiness are simultaneously confirmed. &
SHARD, STPA \\
$R_{17}$ & Allow remote control of end-effectors &
Remote control shall be disabled during exposure and enabled only when posture validity and patient readiness are confirmed. &
SHARD \\
$R_{18}$ & Provide feedback to radiographer and patient &
System feedback shall explicitly indicate system state, pending checks, and reasons for delay using multimodal cues where appropriate. &
STPA \\
$R_{19}$ & Initiate patient-centric HRI strategies &
Patient-centric communication shall be triggered during delays, pauses, or detected discomfort to maintain reassurance and understanding. &
STPA \\
\bottomrule
\end{tabular}
\label{tab:refined_requirements}
\end{table*}

For example, requirements related to motion and exposure (e.g., $R_{4}$ for posture assistance, and $R_{16}$ for imaging accuracy) were refined to include explicit gating on posture stability ($R_{4}$) and patient readiness and arm immobility ($R_{16}$), reflecting high-risk sequencing deviations identified through SHARD. Additional constraints on workflow sequencing and validation were introduced through requirements such as $R_{1}$ (command execution gating) and $R_{6}$ (revalidation following posture change). In contrast, requirements concerning patient wellbeing, feedback, and interaction (e.g., $R_{10}$, $R_{18}$, and $R_{19}$) were strengthened to address unsafe human actions identified through STPA, such as premature confirmations ($R_{10}$) or delayed responses to discomfort ($R_{19}$). In this way, the table demonstrates how abstract safety findings were systematically translated into concrete, testable requirements that constrain system behaviour and reduce reliance on correct human timing or interpretation alone.

While most findings from the SHARD and STPA analyses were incorporated as refinements of the existing system requirements, a small number of additional requirements emerged that could not be expressed as modifications of individual functions alone. Table~\ref{tab:new_requirements} summarises these newly introduced requirements. Requirements such as $R_{20}$ and $R_{21}$ address systematic risks related to premature action and unstable system states by enforcing confirmation structure and stabilisation periods. Others, such as $R_{22}$ and $R_{26}$, mitigate hazards arising from human mental-model mismatch and ambiguous control authority by improving automation transparency and role clarity. Stage-specific requirements (e.g., $R_{23}$–$R_{25}$) focus on safe recovery, revalidation, and exposure readiness following interruptions or faults. 

A representative example of how the analyses informed the refinement process can be seen in the handling of robotic arm motion during patient positioning. Earlier requirements specified that the system should support smooth arm movement and allow radiographer control (e.g., $R_{7}$ and $R_{10}$), but the safety analyses revealed that these capabilities alone were insufficient to prevent hazardous situations arising from premature motion, interruptions, or patient movement. As a result, existing requirements were refined to include explicit gating conditions, such as disabling arm motion unless posture validity and patient readiness are confirmed, while new safety requirements were introduced to address gaps identified by the analyses. In particular, requirement $R_{21}$ enforces a stabilisation period following posture changes, requirement $R_{23}$ mandates revalidation after interruptions or detected movement, and requirement $R_{24}$ constrains motion and exposure until all safety-critical conditions are satisfied. Together, these refined and newly introduced rules illustrate how functional capabilities are augmented with safety constraints to ensure that robotic assistance remains responsive while preventing unsafe sequencing and over-reliance on user actions.

\begin{table}[!t]
\centering
\caption{Additional safety requirements identified through the SHARD and STPA hazard analyses.}
\sffamily
\fontsize{7.3}{8.1}\selectfont
\renewcommand{\arraystretch}{0.85}
\rowcolors{2}{gray!10}{white}
\begin{tabular}{l p{6cm} p{1cm}}
\toprule
\textbf{ID} & \textbf{Description} & \textbf{Method.} \\
\midrule

$R_{20}$ & Safety-critical actions (e.g., motion initiation, X-ray exposure, patient release) shall not be enabled through a single user action alone, but shall require multi-step or multi-source confirmation. & SHARD \\
$R_{21}$ & The system shall enforce a minimum stabilisation period following posture changes, interruptions, or re-positioning before allowing motion planning or X-ray exposure. & SHARD \\
$R_{22}$ & The system shall clearly indicate when automated decisions (e.g., posture detection or trajectory validation) are provisional, unstable, or awaiting confirmation. & STPA \\
$R_{23}$ & Following any interruption, detected fault, or unexpected patient movement, the system shall require revalidation of posture, trajectory, and patient readiness before resuming motion or exposure. & SHARD, STPA \\
$R_{24}$ & X-ray exposure shall remain locked until posture stability, arm immobility, patient assent, and radiographer confirmation are simultaneously satisfied. & SHARD, STPA \\
$R_{25}$ & Upon fault detection, interruption, or session abandonment, the system shall transition the robot to a predefined low-force, low-rigidity safe posture prior to user intervention. & SHARD \\
$R_{26}$ & At each stage of the process, responsibility for progression (system-driven or radiographer-driven) shall be explicit and visible to prevent ambiguity in control authority. & STPA \\
$R_{27}$ & Critical patient-facing instructions shall require explicit acknowledgement or confirmation of understanding before progression to the next safety-critical step. & STPA \\

\bottomrule
\end{tabular}
\label{tab:new_requirements}
\end{table}

\section{Discussion \label{sec:discussion}}

This section discusses the implications of the process modelling and safety analyses that underpin the methodology presented in this paper, focusing on what they reveal about the design of assistive robotic systems for medical applications. It reflects on how the combined use of SHARD and STPA can shape understanding of risk, inform design refinement, and expose broader challenges related to human–robot interaction, clinical workflow integration, and patient experience. In doing so, it situates the MammoBot case study within wider considerations of safety, usability, and ethical deployment of robotic assistance in healthcare, highlighting both the strengths of the proposed approach and the limitations that motivate future work.

\paragraph{Integrating process modelling with SHARD and STPA}
This work demonstrates the value of combining process modelling with complementary safety analysis techniques to reason about complex human–robot interaction in a clinical setting. The activity diagram provided a shared and explicit representation of the MammoBot-assisted breast screening workflow, grounding subsequent analyses in a concrete task structure rather than abstract system descriptions. Building on this model, SHARD and STPA were applied to examine safety from different but mutually reinforcing perspectives. SHARD enabled a systematic exploration of how technical and procedural deviations could arise at each step of the process, while STPA focused attention on unsafe interactions between human users and automated system behaviour. The combination of these methods revealed hazards that would likely have remained obscured if either technique had been applied in isolation.

\paragraph{Insights into human–robot interaction and clinical workflows}
A key insight from the analyses is that many safety risks arise not from component failures, but from predictable and reasonable human behaviours occurring at unsafe moments. The STPA results, in particular, highlight timing-related issues, premature confirmations, misinterpretation of system state, and over-reliance on automation as recurring contributors to hazardous situations. Importantly, the analysis distinguishes between radiographer and patient roles rather than treating “the user” as a homogeneous entity. Radiographer-related risks often involve interpretation, attention, and sequencing of actions, whereas patient-related risks are more closely linked to movement, discomfort, fatigue, or limited ability to communicate. This distinction underscores the need for role-specific safeguards and reinforces the importance of designing robotic assistance systems that accommodate variability in human capability and behaviour rather than assuming ideal interaction.

\paragraph{Non-physical harm}
A further goal, with respect to the unique challenge of robot integration with vulnerable patients in clinical settings, is to anticipate how unintended actions could risk non-physical harm. Adaptation to non-physical forms of harm involves examining conditions that have potentially adverse effects on individual patients and their well-being. These are actions that lead to psychological and/or socio-ethical impact of  severity levels ranging from ‘mild annoyance’ and ‘concern or anxiety’ to ‘deep distress’ from robot interaction.  These hazards or harms emerge from the interrelationship between the robot and the patient’s real-world, lived experience.

In Tables~\ref{tab:SHARD_table} and~\ref{tab:UCA_CUE_table}, we follow a hazard severity level that includes only one category of non-severe, non-physical harm (``Annoyance''). However, accounting for non-physical harm, in this context, spans degrees of severity and can range from having slight psychological, social, ethical, and cultural effect to more serious violations impacting democratic and pro-social values such as respect for patient autonomy, privacy, and human dignity. Even seemingly small or inconsequential forms of non-physical harm can be aggregated across the robot interaction escalating rapidly into far greater, highly consequential harm. Consider how mild annoyance, if left unaddressed, can quickly escalate to intense frustration and psychological distress.  Accordingly, with the assistance of end-users and clinicians, non-physical harm can be expanded in the hazard analysis to include a broader range of non-physical impact with categorised increasing harm severity, necessitating further harm-reduction mitigation. Beyond individual interaction events, assistive robotic systems such as MammoBot must also be understood within a broader socio-technical context in which safety is shaped not only by technical reliability but also by social, ethical, and cultural dynamics~\cite{shelby2023sociotechnical}.

\paragraph{Exploitation}
While the methodology provides a rigorous basis for design refinement, several limitations should be acknowledged. The results are derived from a modelled workflow and expert-informed assumptions rather than from a deployed clinical system. Hazard severity assessments are qualitative, and the effectiveness of the proposed mitigations depends on reliable sensing, user interface design, and system integration. The refined requirements should therefore be viewed as inputs to subsequent verification and validation activities rather than as guarantees of safety. Future work should include implementation-level verification to ensure that gating logic, interlocks, and timing constraints are correctly enforced, as well as empirical validation through usability testing and simulated or clinical evaluations to assess how users interact with the refined system under realistic conditions.

\section{Conclusion and Future Work \label{sec:conclusion}}

This paper has presented a systematic hazard management study for MammoBot, an embodied assistive robotic system aimed at improving access to breast screening for patients with physical impairments. By integrating process modelling with SHARD and STPA, the work demonstrates how safety considerations can be embedded early in the design of human–robot collaborative systems operating in sensitive clinical contexts.

The use of an activity diagram provided a concrete and shared representation of the breast screening workflow, enabling hazards to be analysed in relation to realistic tasks, decision points, and human–robot interactions. SHARD revealed how technical and procedural deviations—particularly those related to timing, validation, and incorrect values—could lead to unsafe system states. STPA complemented this by exposing how predictable human behaviours, such as premature confirmation, delayed responses to discomfort, or misinterpretation of system feedback, can interact with automation in hazardous ways even when individual components function correctly.

A key outcome of these analyses is the refinement of the system requirements, where abstract functional goals were strengthened with explicit safety constraints, including gating logic, stabilisation periods, revalidation after interruptions, and non-bypassable safety checks. A small set of additional requirements was also introduced to address cross-cutting risks that could not be mitigated through adjusting the original functional refinements alone. Together, these requirements provide a traceable link between identified hazards and concrete design constraints, supporting safer human–robot collaboration without assuming ideal user behaviour.

The limitations mentioned in Section~\ref{sec:discussion} will inform the next steps we planned for advancing our hazard management methodology. In particular, the methodology is based on a modelled workflow and expert input rather than observations from a deployed clinical system, and hazard severity assessments are qualitative. Consequently, the results should be viewed as inputs to subsequent design, verification, and validation activities rather than as evidence of assured safety. Our future work will focus on implementation-level verification of the refined requirements, including testing as well as empirical evaluation through usability studies, simulation, and eventually through clinical trials. Finally, in further extensions we will also consider richer representations of non-physical harm and ethical impacts, ensuring that embodied assistive systems such as MammoBot remain safe, trustworthy, and acceptable to both patients and clinicians.

\section*{Acknowledgements}
This work was supported by Cancer Research UK grant EDDPMA-Nov23/100024 and Breast Cancer Now grant 2025.04.IPHEW1812. The authors are grateful to the Centre for Assuring Autonomy at the University of York for providing the research environment that enabled this work. We also thank York's Institute for Safe Autonomy for providing facilities to establish the MammoBot testbed and for hosting meetings with project partners, clinicians, and other stakeholders.

\bibliographystyle{elsarticle-num}
\bibliography{references}

@Article{robotics10020065,
AUTHOR = {Valori, Marcello and Scibilia, Adriano and Fassi, Irene and Saenz, José and Behrens, Roland and Herbster, Sebastian and Bidard, Catherine and Lucet, Eric and Magisson, Alice and Schaake, Leendert and Bessler, Jule and Prange-Lasonder, Gerdienke B. and Kühnrich, Morten and Lassen, Aske B. and Nielsen, Kurt},
TITLE = {Validating Safety in Human–Robot Collaboration: Standards and New Perspectives},
JOURNAL = {Robotics},
VOLUME = {10},
YEAR = {2021},
NUMBER = {2},
ARTICLE-NUMBER = {65},
DOI = {10.3390/robotics10020065}
}

@inproceedings{shelby2023sociotechnical,
  title={Sociotechnical harms of algorithmic systems: Scoping a taxonomy for harm reduction},
  author={Shelby, Renee and Rismani, Shalaleh and Henne, Kathryn and Moon, AJung and Rostamzadeh, Negar and Nicholas, Paul and Yilla-Akbari, N'Mah and Gallegos, Jess and Smart, Andrew and Garcia, Emilio and others},
  booktitle={Proceedings of the 2023 AAAI/ACM Conference on AI, Ethics, and Society},
  pages={723--741},
  year={2023}
}

@ARTICLE{10.3389/frobt.2021.667316,
AUTHOR={Delgado Bellamy, Daniel  and Chance, Gregory  and Caleb-Solly, Praminda  and Dogramadzi, Sanja },
TITLE={Safety Assessment Review of a Dressing Assistance Robot},
JOURNAL={Frontiers in Robotics and AI},
VOLUME={8},
YEAR={2021},
DOI={10.3389/frobt.2021.667316}
}

@article{GUIOCHET2016225,
title = {Hazard analysis of human–robot interactions with HAZOP–UML},
journal = {Safety Science},
volume = {84},
pages = {225-237},
year = {2016},
doi = {https://doi.org/10.1016/j.ssci.2015.12.017},
author = {Jérémie Guiochet}
}

@article{CHEMWENO2020104832,
title = {Orienting safety assurance with outcomes of hazard analysis and risk assessment: A review of the ISO 15066 standard for collaborative robot systems},
journal = {Safety Science},
volume = {129},
pages = {104832},
year = {2020},
doi = {https://doi.org/10.1016/j.ssci.2020.104832},
author = {Peter Chemweno and Liliane Pintelon and Wilm Decre}
}

@book{crawley2015hazop,
  title={HAZOP: Guide to best practice},
  author={Crawley, Frank and Tyler, Brian},
  year={2015},
  publisher={Elsevier}
}

@book{kletz2018hazop,
  title={Hazop \& Hazan: identifying and assessing process industry hazards},
  author={Kletz, Trevor A},
  year={2018},
  publisher={CRC Press}
}

@phdthesis{pumfrey1999principled,
  title={The principled design of computer system safety analyses.},
  author={Pumfrey, David John},
  year={1999},
  school={University of York}
}

@article{10.1145/381766.381770,
    author = {Fenelon, P. and McDermid, J. A. and Nicolson, M. and Pumfrey, D. J.},
    title = {Towards integrated safety analysis and design},
    year = {1994},
    publisher = {Association for Computing Machinery},
    volume = {2},
    number = {1},
    doi = {10.1145/381766.381770},
    journal = {SIGAPP Applied Computing Review},
    pages = {21–32}
}

@INPROCEEDINGS{521885,
  author={McDermid, J.A. and Nicholson, M. and Pumfrey, D.J. and Fenelon, P.},
  booktitle={Proceedings of the 10th Annual Conference on Computer Assurance Systems Integrity, Software Safety and Process Security'}, 
  title={Experience with the application of HAZOP to computer-based systems}, 
  year={1995},
  volume={},
  number={},
  pages={37-48},
  doi={10.1109/CMPASS.1995.521885}
}

@article{Marmot2013,
  author  = {Marmot, M. G. and Altman, D. G. and Cameron, D. A. and Dewar, J. A. and Thompson, S. G. and Wilcox, M.},
  title   = {The benefits and harms of breast cancer screening: an independent review},
  journal = {British Journal of Cancer},
  year    = {2013},
  volume  = {108},
  number  = {11},
  pages   = {2205--2240}
}

@misc{NYBSS2023,
  author      = {{North Yorkshire Breast Screening Service}},
  title       = {NBSS report: 3-year cycle of clients},
  year        = {2023},
  note        = {Report covers the period from 2020 May 1 to 2023 May 31}
}

@misc{PHE2023guidance,
  author       = {{Public Health England}},
  title        = {Guidance for breast screening mammographers},
  organization = {GOV.UK},
  year         = {2023},
  url          = {https://www.gov.uk/government/publications/breast-screening-quality-assurance-for-mammography-and-radiography/guidance-for-breast-screening-mammographers},
  urldate      = {2024-11-28}
}

@article{sung2021global,
  title={Global cancer statistics 2020: {GLOBOCAN} estimates of incidence and mortality worldwide for 36 cancers in 185 countries},
  author={Sung, Hyuna and Ferlay, Jacques and Siegel, Rebecca L and Laversanne, Mathieu and Soerjomataram, Isabelle and Jemal, Ahmedin and Bray, Freddie},
  journal={CA: a cancer journal for clinicians},
  volume={71},
  number={3},
  pages={209--249},
  year={2021},
  publisher={Wiley Online Library}
}

@misc{NHS2021,
  author       = {{National Health Service}},
  title        = {When you'll be invited for breast screening and who should go},
  year         = {2021},
  url          = {https://www.nhs.uk/tests-and-treatments/breast-screening-mammogram/when-youll-be-invited-and-who-should-go/},
  urldate      = {2025-10-12},
  note         = {Page last reviewed: 10 September 2021}
}

@misc{PHE2018inequalities,
  author       = {{Public Health England}},
  title        = {Breast screening: reducing inequalities},
  organization = {GOV.UK},
  year         = {2018},
  url          = {https://www.gov.uk/government/publications/breast-screening-identifying-and-reducing-inequalities/breast-screening-reducing-inequalities},
  urldate      = {2024-11-28}
}

@book{10.7551/mitpress,
    author = {Leveson, Nancy G.},
    title = {Engineering a Safer World: {S}ystems Thinking Applied to Safety},
    publisher = {The MIT Press},
    year = {2012},
    doi = {10.7551/mitpress/8179.001.0001}
}

@article{LEVESON2004237,
title = {A new accident model for engineering safer systems},
journal = {Safety Science},
volume = {42},
number = {4},
pages = {237-270},
year = {2004},
doi = {https://doi.org/10.1016/S0925-7535(03)00047-X},
author = {Nancy Leveson}
}

@manual{LevesonThomas2018,
  author       = {Leveson, Nancy G. and Thomas, John P.},
  title        = {STPA Handbook},
  year         = {2018},
  organization = {MIT Partnership for Systems Approaches to Safety and Security (PSASS)},
  url          = {http://psas.scripts.mit.edu/home/materials/},
  note         = {Accessed: 2025-11-25}
}

@article{BOLBOT2019179,
title = {Vulnerabilities and safety assurance methods in Cyber-Physical Systems: A comprehensive review},
journal = {Reliability Engineering \& System Safety},
volume = {182},
pages = {179-193},
year = {2019},
issn = {0951-8320},
doi = {https://doi.org/10.1016/j.ress.2018.09.004},
author = {Victor Bolbot and Gerasimos Theotokatos and Luminita Manuela Bujorianu and Evangelos Boulougouris and Dracos Vassalos}
}

@inbook{doi:https://doi.org/10.1002/9781118974339.ch7,
author = {Dixon, Jack},
publisher = {John Wiley \& Sons, Ltd},
title = {System Safety Hazard Analysis},
booktitle = {Design for Safety},
chapter = {7},
pages = {125-152},
doi = {https://doi.org/10.1002/9781118974339.ch7},
year = {2018}
}

@article{rooney2004root,
  title={Root cause analysis for beginners},
  author={Rooney, James J and Heuvel, LN Vanden},
  journal={Quality progress},
  volume={37},
  number={7},
  pages={45--56},
  year={2004},
  publisher={ASQC American Society for Quality Control}
}

@ARTICLE{5222114,
  author={Lee, W. S. and Grosh, D. L. and Tillman, F. A. and Lie, C. H.},
  journal={IEEE Transactions on Reliability}, 
  title={Fault Tree Analysis, Methods, and Applications: A Review}, 
  year={1985},
  volume={R-34},
  number={3},
  pages={194-203},
  doi={10.1109/TR.1985.5222114}
}

@book{stamatis2003failure,
  title={Failure mode and effect analysis},
  author={Stamatis, Diomidis H},
  year={2003},
  publisher={Quality Press}
}

@article{DAKWAT2018130,
title = {System safety assessment based on STPA and model checking},
journal = {Safety Science},
volume = {109},
pages = {130-143},
year = {2018},
doi = {https://doi.org/10.1016/j.ssci.2018.05.009},
author = {Alheri Longji Dakwat and Emilia Villani}
}

@article{burkman2017further,
  title={Further development of a robotic-assisted transfer device},
  author={Burkman, Jessica and Grindle, Garrett and Wang, Hongwu and Kelleher, Annmarie and Cooper, Rory A},
  journal={Topics in spinal cord injury rehabilitation},
  volume={23},
  number={2},
  pages={140--146},
  year={2017},
  publisher={Thomas Land Publishers, Inc.}
}

@article{hodkin2018automated,
  title={Automated FES for upper limb rehabilitation following stroke and spinal cord injury},
  author={Hodkin, Edmund F and Lei, Yuming and Humby, Jonathan and Glover, Isabel S and Choudhury, Supriyo and Kumar, Hrishikesh and Perez, Monica A and Rodgers, Helen and Jackson, Andrew},
  journal={IEEE Transactions on Neural Systems and Rehabilitation Engineering},
  volume={26},
  number={5},
  pages={1067--1074},
  year={2018},
  publisher={IEEE}
}

@article{zhu2024you,
  title={Do you need a hand?--a bimanual robotic dressing assistance scheme},
  author={Zhu, Jihong and Gienger, Michael and Franzese, Giovanni and Kober, Jens},
  journal={IEEE Transactions on Robotics},
  volume={40},
  pages={1906--1919},
  year={2024},
  publisher={IEEE}
}

@ARTICLE{X-HAZOP,
  author={Menon, Catherine and Rainer, Austen and Holthaus, Patrick and Moros, Sílvia and Lakatos, Gabriella},
  journal={IEEE Robotics \& Automation Magazine}, 
  title={X-HAZOP: A Family of Techniques for Ethical Hazard Analysis of Assistive Robots}, 
  year={2025},
  volume={32},
  number={4},
  pages={34-41},
  doi={10.1109/MRA.2025.3590612}
}

@article{OGINNI2023106275,
title = {Applying System-Theoretic Process Analysis (STPA)-based methodology supported by Systems Engineering models to a UK rail project},
journal = {Safety Science},
volume = {167},
pages = {106275},
year = {2023},
doi = {https://doi.org/10.1016/j.ssci.2023.106275},
author = {Dapo Oginni and Fanny Camelia and Mikela Chatzimichailidou and Timothy L.J. Ferris}
}

@INPROCEEDINGS{10173978,
  author={Diemert, Simon and Weber, Jens H.},
  booktitle={2023 IEEE/ACM 18th Symposium on Software Engineering for Adaptive and Self-Managing Systems (SEAMS)}, 
  title={Hazard Analysis for Self-Adaptive Systems Using System-Theoretic Process Analysis}, 
  year={2023},
  volume={},
  number={},
  pages={145-156},
  doi={10.1109/SEAMS59076.2023.00028}
}

@InProceedings{SEFM-hazard-analysis,
author="Masci, Paolo
and Zhang, Yi
and Jones, Paul
and Campos, Jos{\'e} C.",
title="A Hazard Analysis Method for Systematic Identification of Safety Requirements for User Interface Software in Medical Devices",
booktitle="Software Engineering and Formal Methods",
year="2017",
publisher="Springer International Publishing",
pages="284--299"
}

@article{VILLANI2018248,
title = {Survey on human–robot collaboration in industrial settings: Safety, intuitive interfaces and applications},
journal = {Mechatronics},
volume = {55},
pages = {248-266},
year = {2018},
doi = {https://doi.org/10.1016/j.mechatronics.2018.02.009},
author = {Valeria Villani and Fabio Pini and Francesco Leali and Cristian Secchi}
}

@inproceedings{DBLP:conf/sefm/StefanakosCDAL22,
  author       = {Ioannis Stefanakos and
                  Radu Calinescu and
                  James A. Douthwaite and
                  Jonathan M. Aitken and
                  James Law},
  title        = {Safety Controller Synthesis for a Mobile Manufacturing Cobot},
  booktitle    = {Software Engineering and Formal Methods - 20th International Conference,
                  {SEFM} 2022, Berlin, Germany, September 26-30, 2022, Proceedings},
  series       = {Lecture Notes in Computer Science},
  volume       = {13550},
  pages        = {271--287},
  publisher    = {Springer},
  year         = {2022},
  doi          = {10.1007/978-3-031-17108-6\_17}
}

@article{DBLP:journals/scp/GleirscherCDLPA22,
  author       = {Mario Gleirscher and
                  Radu Calinescu and
                  James A. Douthwaite and
                  Benjamin Lesage and
                  Colin Paterson and
                  Jonathan M. Aitken and
                  Rob Alexander and
                  James Law},
  title        = {Verified synthesis of optimal safety controllers for human-robot collaboration},
  journal      = {Sci. Comput. Program.},
  volume       = {218},
  pages        = {102809},
  year         = {2022},
  doi          = {10.1016/J.SCICO.2022.102809}
}

@misc{ISO12100_2010,
  author       = {ISO},
  title        = {{ISO 12100:2010 -- Safety of machinery: General principles for design -- Risk assessment and risk reduction}},
  year         = {2010},
  howpublished = {\url{https://www.iso.org/standard/51528.html}},
  address      = {Geneva, Switzerland},
  organization = {International Organization for Standardization}
}

@misc{ISO13849-1_2015,
  author       = {ISO},
  title        = {{ISO 13849-1:2023 -- Safety of machinery: Safety-related parts of control systems, Part 1: General principles for design}},
  year         = {2023},
  howpublished = {\url{https://www.iso.org/standard/73481.html}},
  address      = {Geneva, Switzerland},
  organization = {International Organization for Standardization}
}

@misc{ISO10218-1_2011,
  author       = {ISO},
  title        = {{ISO 10218-1:2025 -- Robotics: Safety requirements., Part 1: Industrial robots}},
  year         = {2025},
  howpublished = {\url{https://www.iso.org/standard/73933.html}},
  address      = {Geneva, Switzerland},
  organization = {International Organization for Standardization}
}

@misc{ISO15066_2016,
  author       = {ISO},
  title        = {{ISO/TS 15066:2016 -- Robots and robotic devices: Collaborative robots}},
  year         = {2016},
  howpublished = {\url{https://www.iso.org/standard/62996.html}},
  address      = {Geneva, Switzerland},
  organization = {International Organization for Standardization}
}

@article{DBLP:journals/arobots/AjoudaniZIAKK18,
  author       = {Arash Ajoudani and
                  Andrea Maria Zanchettin and
                  Serena Ivaldi and
                  Alin Albu{-}Sch{\"{a}}ffer and
                  Kazuhiro Kosuge and
                  Oussama Khatib},
  title        = {Progress and prospects of the human-robot collaboration},
  journal      = {Auton. Robots},
  volume       = {42},
  number       = {5},
  pages        = {957--975},
  year         = {2018},
  doi          = {10.1007/S10514-017-9677-2}
}

@article{DBLP:journals/csur/LuckcuckFDDF19,
  author       = {Matt Luckcuck and
                  Marie Farrell and
                  Louise A. Dennis and
                  Clare Dixon and
                  Michael Fisher},
  title        = {Formal Specification and Verification of Autonomous Robotic Systems:
                  {A} Survey},
  journal      = {{ACM} Comput. Surv.},
  volume       = {52},
  number       = {5},
  pages        = {100:1--100:41},
  year         = {2019},
  doi          = {10.1145/3342355}
}

@article{DBLP:journals/ijrr/WebsterWADEFP20,
  author       = {Matt Webster and
                  David G. Western and
                  Dejanira Araiza{-}Illan and
                  Clare Dixon and
                  Kerstin Eder and
                  Michael Fisher and
                  Anthony G. Pipe},
  title        = {A corroborative approach to verification and validation of human-robot
                  teams},
  journal      = {Int. J. Robotics Res.},
  volume       = {39},
  number       = {1},
  year         = {2020},
  doi          = {10.1177/0278364919883338}
}

@article{DBLP:journals/robotics/MathesonMZFR19,
  author       = {Eloise Matheson and
                  Riccardo Minto and
                  Emanuele G. G. Zampieri and
                  Maurizio Faccio and
                  Giulio Rosati},
  title        = {Human-Robot Collaboration in Manufacturing Applications: {A} Review},
  journal      = {Robotics},
  volume       = {8},
  number       = {4},
  pages        = {100},
  year         = {2019},
  doi          = {10.3390/ROBOTICS8040100}
}

@article{DBLP:journals/jim/LiHZP24,
  author       = {Weidong Li and
                  Yudie Hu and
                  Yong Zhou and
                  Duc Truong Pham},
  title        = {Safe human-robot collaboration for industrial settings: a survey},
  journal      = {J. Intell. Manuf.},
  volume       = {35},
  number       = {5},
  pages        = {2235--2261},
  year         = {2024},
  doi          = {10.1007/S10845-023-02159-4}
}

@ARTICLE{7079531,
  author={Zanchettin, Andrea Maria and Ceriani, Nicola Maria and Rocco, Paolo and Ding, Hao and Matthias, Björn},
  journal={IEEE Transactions on Automation Science and Engineering}, 
  title={Safety in human-robot collaborative manufacturing environments: Metrics and control}, 
  year={2016},
  volume={13},
  number={2},
  pages={882-893},
  doi={10.1109/TASE.2015.2412256}
}

@INPROCEEDINGS{1501143,
  author={Feil-Seifer, D. and Mataric, M.J.},
  booktitle={9th International Conference on Rehabilitation Robotics, 2005. ICORR 2005.}, 
  title={Defining socially assistive robotics}, 
  year={2005},
  volume={},
  number={},
  pages={465-468},
  doi={10.1109/ICORR.2005.1501143}
}

@misc{IEC80601-2019,
  author       = {ISO},
  title        = {{IEC 80601-2-78:2019 -- Medical electrical equipment: Part 2-78: Particular requirements for basic safety and essential performance of medical robots for rehabilitation, assessment, compensation or alleviation}},
  year         = {2019},
  howpublished = {\url{https://www.iso.org/standard/68474.html}},
  address      = {Geneva, Switzerland},
  organization = {International Organization for Standardization and International Electrotechnical Commission}
}

@article{doi:10.1177/0018720811417254,
author = {Peter A. Hancock and Deborah R. Billings and Kristin E. Schaefer and Jessie Y. C. Chen and Ewart J. de Visser and Raja Parasuraman},
title ={A Meta-Analysis of Factors Affecting Trust in Human-Robot Interaction},
journal = {Human Factors},
volume = {53},
number = {5},
pages = {517-527},
year = {2011},
doi = {10.1177/0018720811417254}
}

@article{annurev-psych-010416-043958,
   author = "Broadbent, Elizabeth",
   title = "Interactions With Robots: The Truths We Reveal About Ourselves", 
   journal= "Annual Review of Psychology",
   year = "2017",
   volume = "68",
   number = "Volume 68, 2017",
   pages = "627-652",
   doi = "https://doi.org/10.1146/annurev-psych-010416-043958",
   publisher = "Annual Reviews"
  }

@inproceedings{DBLP:journals/corr/abs-2209-14041,
  author       = {Jordan Hamilton and
                  Ioannis Stefanakos and
                  Radu Calinescu and
                  Javier C{\'{a}}mara},
  title        = {Towards Adaptive Planning of Assistive-care Robot Tasks},
  booktitle    = {Proceedings Fourth International Workshop on Formal Methods for Autonomous
                  Systems {(FMAS)} and Fourth International Workshop on Automated and
                  verifiable Software sYstem DEvelopment (ASYDE), FMAS/ASYDE@SEFM 2022,
                  and Fourth International Workshop on Automated and verifiable Software
                  sYstem DEvelopment (ASYDE)},
  series       = {{EPTCS}},
  volume       = {371},
  pages        = {175--183},
  year         = {2022},
  doi          = {10.4204/EPTCS.371.12}
}

@article{Loftis28122018,
author = {Kathryn L. Loftis and Janet Price and Patrick J. Gillich},
title = {Evolution of the Abbreviated Injury Scale: 1990–2015},
journal = {Traffic Injury Prevention},
volume = {19},
number = {sup2},
pages = {S109--S113},
year = {2018},
publisher = {Taylor \& Francis},
doi = {10.1080/15389588.2018.1512747}
}

@book{rumpe2016modeling,
  title={Modeling with UML},
  author={Rumpe, Bernhard},
  volume={98},
  year={2016},
  publisher={Springer}
}

@misc{OMG_UML,
  author       = {{Object Management Group}},
  title        = {Unified Modeling Language (UML) Specification, Version 2.5.1},
  year         = {2017},
  howpublished = {\url{https://www.omg.org/spec/UML/}}
}

@article{https://doi.org/10.1118/1.4811156,
author = {Mahesh, Mahadevappa},
title = {The Essential Physics of Medical Imaging, Third Edition.},
journal = {Medical Physics},
volume = {40},
number = {7},
pages = {077301},
doi = {https://doi.org/10.1118/1.4811156},
year = {2013}
}

@article{doi:10.1177/02783649261420234,
author = {Ioannis Stefanakos and Jordan Hamilton and Radu Calinescu and Javier C\'{a}mara and Thomas Peyrucain and Narc\'{i}s Miguel Banos},
title ={Adaptive planning for assistive-care robotic missions},
journal = {The International Journal of Robotics Research},
pages = {02783649261420234},
year = {2026},
URL = {https://doi.org/10.1177/02783649261420234}
}

@misc{cdc-2025,
  author = {{US National Center for Chronic Disease Prevention and Health Promotion}},
title = {Health and Economic Benefits of Breast Cancer Interventions},
year = {2025},
note = {\url{https://www.cdc.gov/nccdphp/priorities/breast-cancer.html}}
}

@misc{cruk-2024,
  author = {{Cancer Research UK}},
  title = {A helping hand -- a tale of robots, {AI} and accessible breast screening for all},
  year = {2024},
  note = {\url{https://news.cancerresearchuk.org/2024/06/14/a-helping-hand-a-tale-of-robots-ai-and-accessible-breast-screening-for-all/}}
}

@misc{bcn-2026,
  author = {{Breast Cancer Now}},
  title = {Creating a robotic assistant to help everyone access breast screening},
  year = {2025},
  note = {\url{https://breastcancernow.org/our-research/research-centres-and-projects/individual-research-projects/creating-a-robotic-assistant-to-help-everyone-access-breast-screening}}
}

@misc{zhu_2026_19348891,
  author       = {Zhu, Jihong},
  title        = {Robot demonstration video for {MammoBot} Project},
  year         = 2026,
  publisher    = {Zenodo},
  note          = {\url{https://doi.org/10.5281/zenodo.19348891}}
}

\onecolumn

\appendix
\section{SHARD Analysis Supplementary Material}
\setcounter{table}{0}
\label{app1}
\RaggedRight
\sffamily
{\fontsize{7.3}{8.1}\selectfont
\begin{tabularx}{\textwidth}{%
  p{1.4cm}p{1.5cm}XXXXXp{1cm}%
}
\caption{SHARD analysis of the MammoBot process. Each row corresponds to a node in the activity diagram (Figure~\ref{fig:process_diagram}) analysed using the SHARD guide words (Omission, Commission, Early, Late, and Value). For each applicable deviation, the table identifies possible causes, potential effects, detection or protection mechanisms, and design recommendations, together with a qualitative hazard level.\label{tab:SHARD_table}} \\

\toprule
\textbf{Node} & \textbf{Guide word} & \textbf{Deviation} & \textbf{Possible Causes} & 
\textbf{Effects} & \textbf{Detection and Protection} & 
\textbf{Justification or Design Recommendation} & \textbf{Hazard Level} \\
\midrule
\endfirsthead

\multicolumn{8}{c}%
{{\bfseries Table \thetable\ (continued)}}\\
\toprule
\textbf{Node} & \textbf{Guide word} & \textbf{Deviation} & \textbf{Possible Causes} & 
\textbf{Effects} & \textbf{Detection and Protection} & 
\textbf{Justification or Design Recommendation} & \textbf{Hazard Level} \\
\midrule
\endhead

\midrule\multicolumn{8}{r}{\small Continued on next page}\\
\bottomrule
\endfoot

\bottomrule
\endlastfoot

\rowcolor{gray!10}System initialisation & Omission & Initialisation not executed & Boot failure; skipped by operator & System remains in undefined/unverified state; safety interlocks and configurations not guaranteed active & Self-test sequence mandatory & Block any operation until initialisation passes & High \\
& Commission & Initialisation repeated or wrong profile loaded & Restart timing conflict between modules; wrong configuration loaded & System configuration resets or becomes inconsistent with current session context & Lock initialisation during procedure & Prevent re-initialisation unless idle & Medium \\
\rowcolor{gray!10}& Early & Initialisation completed before all devices/services ready & Dependency race between subsystems; sensors not yet stabilised & System enters operational state without all subsystems verified ready & Initialisation routine checks device health; block until sensors signal ready & Stage initialisation with dependency checks; show progress to operator & Medium \\
& Late & Initialisation takes too long & Slow I/O response; failing hardware & System not ready for use; workflow delayed before session start & Initialisation watchdog timer; progress bar visible to operator & Bound initialisation duration; pre-initialise before patient enters room & Annoyance \\
\rowcolor{gray!10}& Value & Incorrect initialisation parameters loaded (calibration, safety limits, coordinate frames) & Corrupt memory; wrong configuration file chosen & System begins operation with inconsistent configuration state; safety limits and reference settings may be incorrect & Cyclic redundancy and plausibility checks; operator alerts on mismatch & Maintain verified ``golden'' calibration set; require calibration sign-off & High \\
\midrule
System ready? & Omission & Readiness check skipped & Control flow bypass; initialisation verification not executed & System proceeds without confirmed safe initial state; subsystems may be unverified & Mandatory system self-check; readiness status indicator & Block progression until system readiness is confirmed & High \\
\rowcolor{gray!10}& Commission & System incorrectly marked ready & Faulty status reporting; software bug; incorrect readiness flag & Workflow progresses despite incomplete or failed initialisation & Cross-check readiness across subsystems; consistency checks & Require agreement of multiple subsystem readiness signals before proceeding & High \\
& Early & Readiness confirmed prematurely & Initialisation sequence incomplete; dependencies not yet stabilised & System enters operation before all subsystems are fully ready & Readiness gated on completion of all initialisation steps & Enforce dependency checks; delay readiness until all subsystems report stable & High \\
\rowcolor{gray!10}& Late & Readiness confirmed too late & Slow system response; delayed status update & Workflow start delayed; unnecessary waiting & Readiness status monitoring; operator feedback & Optimise initialisation timing; provide clear progress indication & Annoyance \\
& Value & Incorrect readiness status (false ready/not ready) & Misreported system state; configuration mismatch; sensor/status error & System may proceed unsafely or be unnecessarily blocked & Plausibility checks on readiness state; diagnostic feedback & Validate readiness logic; ensure accurate state reporting across modules & High \\
\midrule
\rowcolor{gray!10}Identify process stage & Omission & Stage not identified & State estimator fails; inputs unavailable (e.g., image index, arm pose) & Current workflow stage remains undefined; system cannot determine correct next step & Prompt operator; offer manual stage picker with clear options & Provide manual fallback; log cause; minimise reliance on brittle signals & Low \\
& Commission & Wrong stage identified & Out-of-order events; stale context after retry/restart; clock skew; partial sensor updates & System state misaligned with actual workflow phase & Consistency checks vs context (arm pose, interlocks, patient-OK flag, planned image number) & Commit stage only when context is consistent; require operator confirm on discrepancies & Medium \\
\rowcolor{gray!10}& Early & Stage classified before sufficient contextual evidence has established & Too-short observation window; estimator commits on transient cues & Stage state recorded prematurely based on incomplete context & ``Estimating'' state until K consistent cycles; hold-time before commit & Enforce minimum evidence window; show pending status on UI & Medium \\
& Late & Stage identified late & Low scheduling priority; queue backlog & System remains in previous stage state longer than appropriate & Timing alerts; highlight ``waiting for stage'' on UI & Prioritise stage estimation; simplify rules to reduce latency & Annoyance \\
\rowcolor{gray!10}& Value & Incorrect stage attributes (e.g., side/image index, variant) & Metadata mismatch; naming/units mix-up & Stage context stored incorrectly for current session & Schema validation; display stage summary for operator check & Strong typing for stage context; bind to current case/session data & Medium \\
\midrule
Process stage identified? & Omission & Decision not evaluated & Control flow skip; node bypass & Workflow freezes at this gateway & Watchdog to ensure decision evaluated & Guarantee gateway executes each cycle; provide safe default branch & Low \\
\rowcolor{gray!10}& Commission & Decision flips or returns “identified” when not stable & No debounce; noisy inputs; transient context & Thrashing between branches; confusing operator prompts & Require K consecutive consistent evaluations; debounce/hysteresis & Latch decision for a hold-time before allowing another change & Annoyance \\
\midrule
Determine patient posture & Omission & Posture not detected & Camera blocked or failure; algorithm failure & Patient posture state remains unknown to system & Sensor health monitoring; operator notified & Radiographer can override and adjust posture manually & Annoyance \\
\rowcolor{gray!10}& Commission & Wrong posture detected & Misclassification; occlusion; poor lighting & System records an incorrect posture state for the patient & Cross-check with multiple sensors; radiographer review & Combine infrared and depth sensing with operator confirmation; improve detection robustness & High \\
& Early & Posture estimated before patient/sensors stabilise & Patient still moving into position; camera warm-up settling & Posture state recorded from unstable observations & Require stable posture confidence over K frames; sensor “ready” flag & Gate posture estimation on stability/quality checks; show “stabilising” status & Medium \\
\rowcolor{gray!10}& Late & Posture detected too late & Processing lag; system overload & Delay in establishing patient posture state & Response-time monitoring; operator prompt & Allocate compute resources; prioritise posture detection & Annoyance \\
& Value & Wrong posture parameters (angles/heights) & Calibration drift; scaling error; occlusion & Incorrect posture values stored in system state & Plausibility bounds; anomaly detection; radiographer double-check & Regular calibration; robust posture model & High \\
\midrule
\rowcolor{gray!10}Posture detected? & Omission & Check not performed & Logic bypass; software bug & Workflow halts, no trajectory planning possible & Watchdog to ensure decision always executed & Guarantee posture detection check before planning & Low \\
& Commission & Spurious positive (detection flagged when posture not actually acquired) & Noise; mis-trigger from sensor; poor debounce & System progresses with invalid input, leading to unsafe or pointless planning & Cross-check detection with signal quality; require radiographer confirmation & Require stable detection over multiple frames & Medium \\
\midrule
\rowcolor{gray!10}Trajectory planning & Omission & No motion plan produced & Planning algorithm fails; constraints invalid & No trajectory state generated for current posture configuration & Planner retry; fallback to default safe posture & Provide operator option to manually guide arms & Medium \\
& Commission & Plan generated when it should be blocked & Preconditions bypassed; stale posture input used & Trajectory state created despite unmet planning conditions & Pre-checks on plan validity (collision, range, dose) & Require all safety conditions green before planning allowed & High \\
\rowcolor{gray!10}& Early & Plan computed too soon & Triggered before posture stabilised; using provisional data & Trajectory based on incomplete or unstable input data & Input freshness stamps; block planning until inputs stable & Delay planning until validated posture available & High \\
& Late & Plan produced too late & Heavy computation; solver timeout & Delay in establishing movement plan state & Planning time guard; notify operator if slow & Use bounded-time planners; cache common movements & Annoyance \\
\rowcolor{gray!10}& Value & Plan parameters wrong (angles, units, scales) & Calibration/units mismatch; frame of reference error & Generated trajectory contains incorrect parameter values & Simulation check; operator preview of motion path & Strict validation of coordinate frames; enforce consistent units & High \\
\midrule
Trajectory valid? & Omission & Trajectory validity not checked & Validator step skipped due to software error & Unsafe plan executed & Mandatory safety check before motion; interlock & Ensure validation is non-bypassable & High \\
\rowcolor{gray!10}& Commission & Invalid trajectory marked valid & Validation bug; incomplete safety constraints & Collision or overextension risk & Independent secondary validation module & Two independent validation checks for diversity & High \\
& Value & Validation thresholds misconfigured & Margins too tight or too loose; misconfigured safety limits & Either overly cautious (delays) or unsafe clearance & Safety margin monitors; operator alerts & Regular review of thresholds; calibrate margins to clinical practice & High \\
\midrule
\rowcolor{gray!10}Perform arm positioning & Omission & Arm motion not executed when expected & Drive disabled; emergency stop active; communication fault & Session stalls; patient held uncomfortably in partial setup & Clear fault indication on UI; operator alerted to reset & Provide guided recovery steps before reattempt & Medium \\
& Commission & Arm motion initiated without intended authorisation & Spurious command; outdated trajectory; operator mis-click & Patient startled or at risk of unintended contact & Motion only allowed after explicit confirmation; safety limits on speed and force & Require operator confirmation before movement; enforce protective zones & High \\
\rowcolor{gray!10}& Early & Motion begins before all checks are complete & Patient readiness not confirmed; posture validation pending & Movement occurs while patient unprepared, leading to safety/comfort risk & System blocks motion until all prerequisites are confirmed & Link motor enable strictly to patient-OK, posture valid, and operator consent & High \\
& Late & Motion starts later than expected & Processing or communication delays; resource contention & Patient discomfort increases; workflow slowed & Monitor execution time; notify operator of unusual delay & Prioritise arm commands; keep motion plans efficient & Annoyance \\
\rowcolor{gray!10}& Value & Wrong motion parameters used (e.g., angles, speed, limits) & Calibration error; unit mismatch; incorrect configuration & Risk of excessive force, overreach, or misalignment & Automatic stop if abnormal force/position detected; plausibility checks & Regular calibration; conservative default limits; operator preview before execution & High \\
\midrule
Fault detected? & Omission & Real fault not detected & Fault sensor disabled or failed; alarm muted; check not running & Unsafe condition persists (e.g., obstruction or controller failure) & Independent hardware interlocks; periodic self-tests; heartbeat monitoring & Treat fault monitoring as mandatory; block motion if monitors are offline & High \\
\rowcolor{gray!10}& Commission & Fault reported when none exists & Noisy signal; thresholds too tight; brief power or network interruption & Unnecessary stop; workflow disruption; patient confidence affected & Basic plausibility checks; short re-check before latching; event log & Add debouncing; classify as ``warning'' first when uncertain; require operator confirmation to resume & Annoyance \\
& Early & Fault latched on a short transient event & Momentary sensor glitch; transient spike; overly aggressive threshold & Unnecessary stop mid-motion & Require persistence over several samples before latching & Set minimum duration before a fault triggers a stop & Low \\
\rowcolor{gray!10}& Late & Fault detected after a delay & Slow polling; aggressive filtering & Motion/exposure continues longer than safe & Fast, always-available stop path; maximum detection time budget & Prioritise fault checks; keep detection fast on safety-critical channels & High \\
& Value & Wrong fault type or severity shown & Mislabelled codes; incorrect mapping or thresholds & Over- or under-reaction; poor recovery guidance & Cross-check fault against context (pose, speed, current); clear on-screen text & Standardise codes and severities; show actionable recovery steps & Medium \\
\midrule
\rowcolor{gray!10}HRI interruption? 
& Omission 
& Protective stop request not recognised 
& Speech recognition failure; UI input missed; manual protective-stop control unavailable; software error 
& Robot continues motion despite patient/radiographer requesting stop; protective stop not initiated 
& Command acknowledgement feedback (audio/visual); watchdog on stop-input channels; periodic self-test of stop interfaces 
& Provide redundant stop channels (voice, UI, and manual protective-stop control); ensure at least one stop path is safety-rated and independent of UI 
& High \\
& Commission 
& Protective stop triggered unintentionally 
& False positive speech recognition; accidental activation of manual stop; UI mis-click 
& Robot enters protective stop unnecessarily; workflow interruption and possible patient concern 
& Clear stop-state annunciation; event logging 
& Use robust keyword detection and confidence filtering; require deliberate manual action (guarded button/press-and-hold); require explicit confirmation to resume 
& Low \\
\rowcolor{gray!10}& Late 
& Protective stop processed too slowly 
& Speech processing latency; UI thread stall; delayed propagation to motion controller; system overload 
& Motion continues longer than intended before stopping; increased safety risk 
& Stop-response time monitoring; safety controller supervision 
& Route stop requests via a high-priority, safety-rated channel with bounded response time; do not depend on non-real-time UI processing 
& High \\
& Value 
& Protective stop applied incompletely or to wrong subsystems 
& Communication fault; incorrect command routing; partial controller coverage; integration error 
& Some actuators continue moving despite stop request; incomplete safety state achieved 
& System-wide stop-state verification; cross-check all actuator enables = false; safety controller monitoring 
& Ensure protective stop propagates to all motion controllers; verify stop coverage in integration tests; maintain independent emergency stop as ultimate fallback 
& High \\
\midrule
\rowcolor{gray!10}Patient OK? & Omission & Patient wellbeing not checked & Step skipped; staff oversight & Patient may continue despite distress or pain & UI prompt for operator; optional patient input device & Add mandatory check-in before each new process step & High \\
& Commission & Patient marked OK when not & Misread signals; rushed confirmation & Patient discomfort or harm overlooked & Independent confirmation by operator and patient & Require explicit patient assent for ``OK'' confirmation & High \\
\rowcolor{gray!10}& Early & ``OK'' recorded before patient feedback or signals are stable  & Premature confirmation; vital signs still settling & Distress or discomfort overlooked & Delay check until stable readings and verbal confirmation & Require stable signal window and patient assent before acceptance & High \\
& Late & Check performed too late & Staff distracted; delayed prompt & Distress continues longer before action taken & Timer reminders to prompt operator & Require wellbeing checks at regular intervals & Medium \\
\rowcolor{gray!10}& Value & Wrong threshold for wellbeing (too strict/too lenient) & Poorly tuned monitoring; misinterpretation of signals & Either alarm fatigue or missed patient issues & Regular threshold review with clinical staff & Tailor limits to individual patient needs & Medium \\
\midrule
Adjustments needed? & Omission & Adjustment flag not raised when required & Image quality check fails to detect misalignment & Poor images; potential need for repeat session & Automatic post-capture quality review & Use multiple image quality metrics & Medium \\
\rowcolor{gray!10}& Commission & Adjustment flagged when not needed & Over-sensitive metric; false trigger & Unnecessary extra motions; increased patient burden & Radiographer can override suggestion & Tune sensitivity to balance detection vs false alarms & Low \\
& Early & Adjustment flag raised before image stabilised & Transient data or incomplete processing & Unnecessary corrections suggested & Delay evaluation until image fully available & Require analysis over several frames before flagging & Low \\
\rowcolor{gray!10}& Late & Adjustment flagged too late & Quality check delayed & Extra time; patient kept in position longer & Inline quality analysis after capture & Perform analysis on streaming data when possible & Annoyance \\
& Value & Wrong reason for adjustment indicated & Mislabelled metric; incorrect threshold & Radiographer applies unnecessary or wrong correction & Provide clear, actionable adjustment reason & Periodically validate metrics against clinical outcomes & Medium \\
\midrule
\rowcolor{gray!10}Perform positioning adjustments & Omission & No adjustment applied when needed & Missed ``adjustments needed'' signal; operator skips & Images poor; repeat exposure required & Automatic flag to operator if adjustment skipped & Require confirmation if adjustment ignored & Medium \\
& Commission & Adjustment made when not required & Mis-trigger; operator mis-click or override & Extra patient movement; wasted time & Preview adjustment on screen before applying & Only allow adjustments above a minimum threshold & Low \\
\rowcolor{gray!10}& Early & Adjustment applied before quality analysis stabilised & Operator acts on incomplete data & Incorrect or unnecessary movement; patient discomfort & Show ``analysis in progress'' status until stable & Block adjustments until metrics confirmed & High \\
& Late & Adjustment carried out too late & Processing delay; operator busy & Patient moves before correction; capture invalidated & Monitor timing; alert if delayed & Tighten loop between analysis and adjustment & Annoyance \\
\rowcolor{gray!10}& Value & Adjustment magnitude/direction incorrect & Calibration error; sign/units mismatch & Misalignment; possible collision or discomfort & Motion bounds and safety stops; operator preview & Validate calibration regularly; confirm motion visually before applying & High \\
\midrule
Capture X-ray & Omission & X-ray not triggered & Interlock stuck; tube fault; operator oversight & Missed image; procedure prolonged & Readiness check before trigger; clear error display & Provide retry pathway; show status clearly on UI & Medium \\
\rowcolor{gray!10}& Commission & X-ray triggered unintentionally & Accidental input; spurious signal & Unnecessary radiation dose & Two-step confirmation before exposure & Require operator confirmation and hardware enable before trigger & High \\
& Early & X-ray triggered before patient and arms stable & Readiness interlock failure; incorrect readiness state; gating logic bypassed & Poor image quality; repeat exposure required; unnecessary radiation dose; patient discomfort & Synchronise capture with posture stability and readiness signals & Only allow exposure when posture stability, patient readiness, and radiographer confirmation are verified  & High \\
\rowcolor{gray!10}& Late & X-ray triggered too late & Tube warm-up delay; system lag & Motion blur or patient fatigue; repeat required & Synchronise capture with verified patient stability; provide UI feedback on delay & Enforce timing windows for exposure & Medium \\
& Value & Wrong exposure settings used & Incorrect protocol; calibration error & Poor image quality or unnecessary/excessive radiation dose & Protocol integrity check (checksum); real-time dose monitoring & Lock protocols; display settings to operator before capture & High \\
\midrule
\rowcolor{gray!10}Retake needed? & Omission & Retake not requested when necessary & Quality check fails; thresholds too lenient & Poor diagnostic quality; patient may need recall & Radiographer review of image quality; double-check flags & Use conservative thresholds; escalate borderline cases & High \\
& Commission & Retake requested unnecessarily & Over-sensitive quality metrics; false trigger & Unnecessary repeat exposure & Radiographer review and override if needed & Optimise image quality algorithms to reduce false positives & Medium \\
\rowcolor{gray!10}& Late & Retake decided too late & Quality review delayed until after patient released & Patient recalled later; inconvenience; anxiety & Require quality review before patient release & Inline quality checks before discharge & Annoyance \\
& Value & Wrong reason for retake shown & Mislabelled metric or logging error & Harder to correct issue; may repeat same error & Provide structured reasons for retake in UI & Maintain clear taxonomy of retake reasons & Low \\
\midrule
\rowcolor{gray!10}``Release'' patient & Omission & Release action not executed when required & State transition missed; system remains rigid; operator oversight & Patient remains supported/immobilised longer than necessary; discomfort or fatigue increases & Visual status indicators; operator prompt & Automatically transition to compliant mode when safe conditions met & Annoyance \\
& Commission & Release triggered when not intended & Wrong state transition; premature trigger; operator misinterpretation & Support removed while patient still relies on arm support; risk of instability or loss of balance & State-based gating; posture/stability checks before release & Allow release only when patient stable and motion complete & High \\
\rowcolor{gray!10}& Early & Release occurs before positioning or motion fully stabilised & Premature transition; incorrect sequencing & Patient shifts unexpectedly; posture lost; positioning invalidated & Require posture stability and motion-complete flags & Delay release until system and patient stability confirmed & High \\
& Late & Release delayed after positioning complete & State transition delay; operator distraction & Patient held in supported position longer than needed; discomfort & Timeout reminders; comfort prompts & Auto-release when stable and no further motion planned & Annoyance \\
\rowcolor{gray!10}& Value & Incorrect compliance level applied during release (too rigid/too loose) & Configuration error; wrong mode selected & Either insufficient comfort (too rigid) or insufficient support (too loose) & Mode validation; force/position monitoring & Define safe compliance ranges; verify release mode before activation & Medium \\
\midrule
Process Done? & Omission & Completion condition never satisfied & Image counter not updated; state transition missed & Workflow cannot terminate; patient release blocked & Cross-check image count; operator alert if process stalls & Derive completion strictly from verified image captures & Low \\
\rowcolor{gray!10}& Commission & Process marked complete before all required X-rays are taken & Incorrect image count; session mix-up; state corruption & Missing diagnostic data; patient released prematurely & Verify image count against expected protocol & Assert completion only when all required images are verified & Low \\
\end{tabularx}
}

\newpage
\section{STPA Analysis Supplementary Material}
\setcounter{table}{0}
\label{app2}

\begin{table}[!h]
\sffamily
\setlength{\tabcolsep}{4pt}
\fontsize{7.3}{8.1}\selectfont
\centering
\caption{Breakdown of unsafe control actions (UCAs) and common user errors (CUEs) identified through STPA (Figure~\ref{fig:STPA}). The table summarises potential causes, effects, detection or protection mechanisms, and corresponding design recommendations.}
\rowcolors{2}{gray!10}{white}
\begin{tabular}{l p{3.2cm} p{3.2cm} p{3.2cm} p{3.9cm} p{1.05cm}}
\toprule
\textbf{ID} &
\textbf{Possible Causes} &
\textbf{Effects} &
\textbf{Detection and Protection} &
\textbf{Justification or Design\newline Recommendation} &
\textbf{Hazard Level} \\
\midrule


UCA01 & Alert fatigue; time pressure & System used in unsafe or unknown state & Mandatory self-check; persistent warnings & Block progression until checks pass & High \\

UCA02 & Wrong patient selected; UI similarity & Incorrect parameters applied & Patient ID check; profile summary & Require explicit patient confirmation & High \\

UCA03 & Over-trust in setup; skipped verification & Unsafe motion or exposure enabled & Calibration and interlock status checks & Block use unless verified & High \\

UCA04 & Ambiguous indicators; poor UI clarity & Incorrect operator actions & Redundant state indicators & Use clear, standardised system states & Medium \\

UCA05 & Confusing workflow stages & Wrong imaging or motion sequence & Context consistency checks & Restrict stage selection by context & Medium \\

UCA06 & Rushed workflow; missing checks & Premature transition to next stage & Precondition validation & Enforce completion of required checks & Medium \\

UCA07 & State desynchronisation & Workflow confusion; unsafe action & Cross-check system and UI state & Highlight mismatches explicitly & Medium \\

UCA08 & Over-trust in automation & Motion planned from unstable posture & Confidence thresholds; warnings & Require stable posture confirmation & High \\

UCA09 & Ignored warnings; alert fatigue & Unsafe planning assumptions & Salient low-confidence alerts & Escalate ambiguous detections & High \\

UCA10 & Inattention; delayed feedback & Motion based on outdated posture & Continuous monitoring & Revalidate posture before motion & High \\

UCA11 & Discomfort; misunderstanding & Invalid posture detection & Motion sensors; alerts & Pause if patient moves & Medium \\

UCA12 & Unclear instructions & Incorrect patient stance & Visual/verbal guidance & Use simple, multimodal instructions & Medium \\

UCA13 & Fear; communication barriers & Distress unaddressed & Patient input; monitoring & Provide easy stop/feedback channel & High \\

UCA14 & Incomplete review & Unsafe trajectory approved & Safety margin indicators & Require explicit plan review & High \\

UCA15 & Misread validation results & Unsafe motion execution & Clear validation outcome display & Use binary, unambiguous results & High \\

UCA16 & Stale data reuse & Unsafe or inefficient motion & Data freshness checks & Invalidate outdated plans & High \\

UCA17 & Patient movement & Planning invalidated & Motion detection & Re-plan on movement & Medium \\

UCA18 & Ignored distress signals & Harm or discomfort & Force/voice monitoring & Require patient-ready confirmation & High \\

UCA19 & Over-reliance on automation & Unsafe contact forces & Force thresholds; alarms & Require active monitoring & High \\

UCA20 & Delayed response & Prolonged discomfort & Audible/visual alerts & Escalate unresolved cues & Medium \\

UCA21 & Automation bias & Missed unsafe motion & Redundant sensing & Encourage human oversight & Medium \\

UCA22 & Startle or fear & Collision or misalignment & Motion stop on movement & Pause on unexpected motion & High \\

UCA23 & Communication difficulty & Unsafe continuation & Patient input devices & Provide low-effort stop signal & High \\

UCA24 & Instruction ambiguity & Loss of posture stability & Clarification prompts & Simplify instructions & Medium \\

UCA25 & Over-sensitive metrics & Unnecessary adjustments & Operator override & Require justification for adjustment & Low \\

UCA26 & Focus on task & Discomfort accumulates & Periodic comfort prompts & Enforce comfort checks & Medium \\

UCA27 & Patient reacts to motion & Misalignment & Motion monitoring & Slow, predictable movement & Medium \\

UCA28 & Rushed workflow & Unnecessary radiation & Exposure gating & Require posture stability & High \\

UCA29 & Wrong protocol selected & Poor image or excess dose & Protocol checksum & Lock protocol changes & High \\

UCA30 & Late discomfort report & Exposure not stopped & Voice/emergency stop & Always-active stop channel & High \\

UCA31 & Involuntary movement & Image artefacts & Motion detection & Abort on motion & Medium \\

UCA32 & Poor communication & Patient startled by motion & Verbal/system cue & Announce arm release & Annoyance \\

UCA33 & Skipped quality control & Missed diagnostic issues & Quality control checklist & Block release until quality control complete & High \\

UCA34 & Misunderstood completion & Collision risk & Safe-position confirmation & Enforce safe rest before release & High \\

UCA35 & Workflow misunderstanding & Premature movement & Clear end-of-process cue & Explicit completion signal & Medium \\


CUE01 & Memory lapse; distraction & Required step omitted & Checklists; prompts & Enforce step completion & Medium \\

CUE02 & Ambiguous feedback & Incorrect user response & Redundant feedback & Improve feedback clarity & Annoyance \\

CUE03 & Poor indicator design & Misread system state & Status cross-check & Standardise indicators & Medium \\

CUE04 & Timing mismatch & Premature or delayed action & Temporal guards & Gate actions by state & Medium \\

CUE05 & Automation bias & Missed unsafe condition & Independent monitoring & Require human confirmation & Medium \\

CUE06 & Physical/cognitive limits & Reduced interaction ability & Adaptive interfaces & Provide alternative inputs & Medium \\

CUE07 & Instruction misinterpretation & Incorrect action taken & Clarification prompts & Simplify and repeat instructions & Medium \\

\bottomrule
\end{tabular}
\label{tab:UCA_CUE_table}
\end{table}

\twocolumn

\end{document}